\documentclass[lettersize,journal]{IEEEtran}
\usepackage{amsmath}
\usepackage[capitalize]{cleveref}
\usepackage{subcaption}
\usepackage{booktabs}
\usepackage{multicol}
\usepackage{multirow}
\usepackage[table]{xcolor}
\usepackage{tabularray}
\usepackage{caption}
\usepackage{float}
\usepackage{graphicx}
\usepackage{amsmath,amsfonts}
\usepackage{algorithmic}
\usepackage{algorithm}
\usepackage{array}
\usepackage[caption=false,font=normalsize,labelfont=sf,textfont=sf]{subfig}
\usepackage{textcomp}
\usepackage{stfloats}
\usepackage{url}
\usepackage{verbatim}
\usepackage{graphicx}
\usepackage{cite}
\usepackage[capitalize]{cleveref}
\usepackage{tabularx} 
\usepackage{booktabs} 
\usepackage[table]{xcolor} 
\begin{document}

\title{iGSP:Implicit Gradient Subspace Projection for Efficient Continual Learning of
Vision-Language Models}

\author{Xuezhi~Cui,
        Dongbo~Zhou,~\IEEEmembership{Member,~IEEE,}
        Wang~Guo,
        Zeyuan~Wang,
        Ziyu~Li, 
        Gaozhi~Zhou, 
        Xian~Li,
        Ling~Zhao,
        Wentao~Yang,
        Chao~Tao,~\IEEEmembership{Member,~IEEE,}
        Haifeng~Li,~\IEEEmembership{Senior Member,~IEEE}
\thanks{Xuezhi Cui, Wang Guo, Zeyuan Wang, Ziyu Li, Gaozhi~Zhou, Ling~Zhao, Xian~Li, Chao~Tao and Haifeng Li are with the School of Geosciences and Info-Physics, Central South University, Changsha 410083, China.}
\thanks{Wentao~Yang is with School of Earth Sciences and Spatial Information Engineering, Hunan University of Science and Technology, Xiangtan 411201, China.}
\thanks{Dongbo~Zhou is with the Faculty of Artifi cial Intelligence in Education in Central China Normal University, Wuhan 430079, China.

(Corresponding author: Dongbo~Zhou)

(e-mail: zhoudongbo@ccnu.edu.cn)}}

\markboth{Journal of \LaTeX\ Class Files,~Vol.~14, No.~8, August~2021}%
{Shell \MakeLowercase{\textit{et al.}}: A Sample Article Using IEEEtran.cls for IEEE Journals}


\maketitle

\begin{abstract}
Vision-Language Models require efficient adaptation to continually emerging downstream tasks. While Parameter-Efficient Fine-Tuning mitigates catastrophic forgetting, assigning isolated modules per task leads to parameter explosion. Conversely, recent similarity-driven sharing mechanisms falsely equate superficial visual similarity with underlying alignment consistency. This fundamental mismatch triggers severe negative transfer between visually similar but logically distinct tasks and fails to exploit alignment reuse across visually diverse ones. We argue thatalignment sharing is fundamentally a geometric problem of overlapping optimization trajectories within shared low-rank subspaces. Grounded in this insight, we propose iGSP, a novel framework that achieves efficient adaptation via implicit gradient subspace projection. Leveraging the early convergence of MoE routers to establish the subspace basis, iGSP bifurcates the adaptation process into two phases. First, the Subspace Identification phase introduces candidate experts via basis pre-expansion, applies a novel subspace-constrained regularization to implicitly project new task gradients onto the historical subspace, and precisely prunes redundant dimensions by treating routing probabilities as gradient flow indicators, ultimately to maximize knowledge reuse. Second, the Orthogonal Subspace Fine-Tuning phase fixes this structural basis and removes the regularization to rapidly fit the task-specific residual loss. Extensive experiments on the MTIL benchmark demonstrate that iGSP achieves state-of-the-art accuracy while significantly improving training efficiency, reducing the average trainable parameters by 42.7\% compared to current SOTA methods, and decreasing the final total parameters by 86.9\% relative to counterparts. 
The source code is available at https://github.com/GeoX-Lab/iGSP.
\end{abstract}


\section{Introduction}
\IEEEPARstart{V}{ision}–Language Models (VLMs) trained at scale exhibit strong, general-purpose cross-modal alignment\cite{clip,llava}. Yet as new tasks continually emerge, VLMs still require continual adaptation\cite{clmoe,zhang2024continual}. Continual Learning (CL) offers a principled path to accumulate capabilities without full retraining by preserving prior knowledge while adapting to novel tasks\cite{bilora,yu2025language,kang2025your}. In VLMs, however, “catastrophic forgetting”\cite{catastrophic} manifests as cross-modal alignment drift\cite{continualsurvey}, where learning the alignment best suited for a new task disrupts the alignment established on pretraining data and earlier tasks, hindering multi-task deployment\cite{aligment}.
Classical CL approaches mitigate forgetting from a knowledge-compatibility perspective: (i) \textit{Knowledge distillation} constrains the new model to match the old model on past tasks \cite{lwf,zscl,wu2024image}; (ii) \textit{Regularization} penalizes updates on parameters deemed important to prior tasks \cite{ewc,re1} and (iii) \textit{Replay} stores or synthesizes previous examples for rehearsal \cite{icarl,gift,replay1,squeezing}. While effective to a degree, these strategies often incur substantial compute or storage overhead, limiting their practicality in resource-constrained settings.

\label{sec:intro}
\begin{figure*}[htbp]
  \centering
   \includegraphics[width=0.9\linewidth]{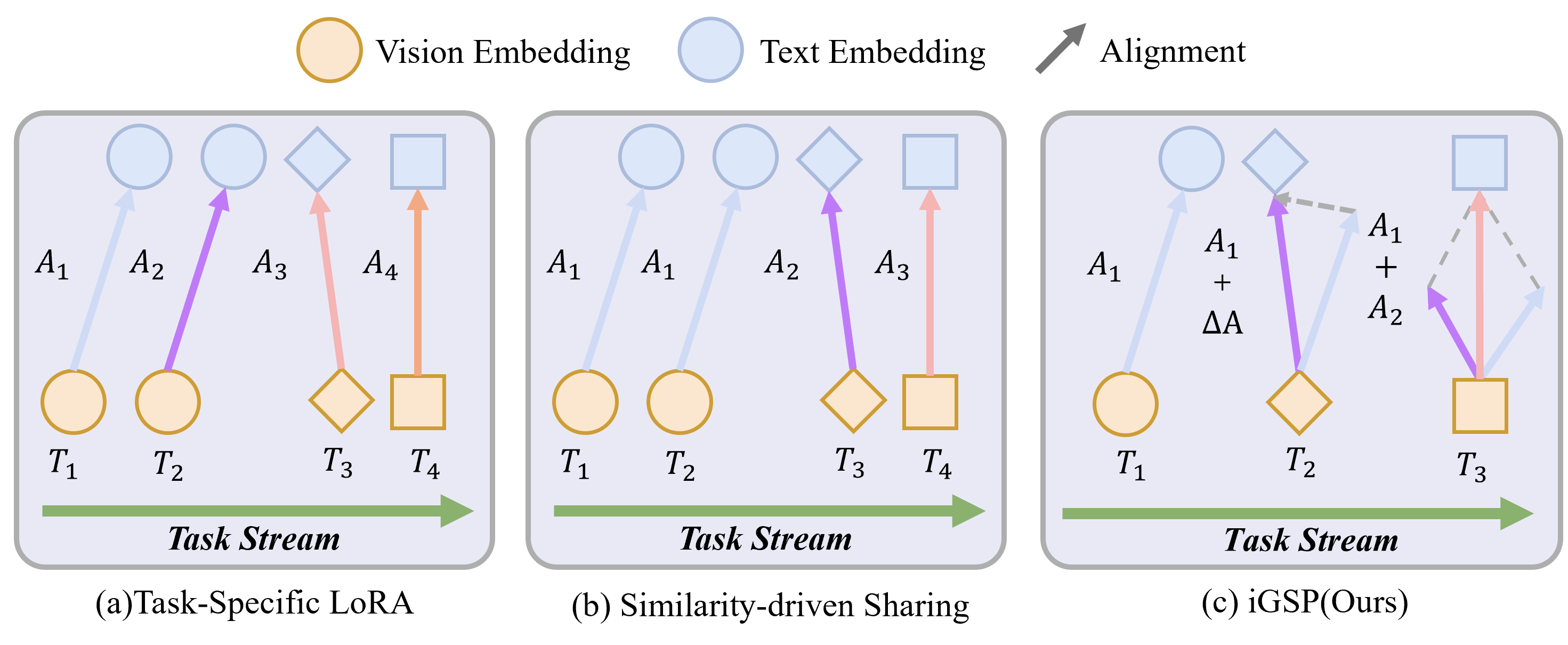}
   \caption{Comparison of alignment strategies in continual learning. (a) Task-specific LoRA: Assigns isolated LoRA modules to each task and precludes any cross-task parameter sharing. (b) Similarity-driven Sharing: Facilitates alignment sharing based strictly on visual proximity where reuse is conditioned on surface-level feature similarity. (c) Ours (iGSP): Employs implicit gradient subspace projection to maximize knowledge reuse across tasks and dynamically introduces task-specific orthogonal experts only to capture residual alignment requirements that lie beyond the capacity of the historical subspace.}
   \label{fig:duibil}
\end{figure*}
 To improve efficiency, recent works such as\cite{xiong2026class} and \cite{l2p} integrate CL with parameter-efficient fine-tuning (PEFT). From a model-expansion perspective, these methods\cite{diki,inflora} freeze the pretrained backbone to retain general representations and attach lightweight, task-wise adapter modules(e.g. LoRA\cite{lora},  Prompts\cite{prefix})to learn task-specific vision-language alignments while minimizing inter-task interference.However, assigning strictly isolated modules to each task inherently ignores the latent shared structure of vision-language alignment. Without a mechanism to discover and reuse these common alignments, the model is forced to independently reconstruct task-specific mappings for every new task, leading to a redundant expansion of the parameter space.

To enable parameter sharing, recent advancements (e.g., \cite{yu2025moe,sema}) introduce similarity-driven mechanisms, utilizing visual embedding proximity as a heuristic to trigger alignment reuse. We argue that this paradigm operates under a fundamentally flawed assumption: falsely equating superficial visual similarity with cross-modal alignment consistency. This misalignment has fatal consequences in Continual Learning. On one hand, visually homogeneous samples might necessitate divergent mapping strategies depending on the task goal; forcing them to share parameters strictly based on appearance triggers severe negative transfer . On the other hand, visually heterogeneous tasks may actually share identical deep-level semantic rules; relying on visual distance completely blinds the model to these critical cross-domain knowledge reuse opportunities, leaving massive parameter redundancy unresolved. We contend that true non-interfering sharing in CL is not dictated by input features, but is intrinsically a geometric problem in the optimization space. Two tasks should share the same alignment basis if and only if their optimization trajectories involve common components that can be projected onto a shared low-rank subspace.

\begin{figure}[htbp]
  \centering
   \includegraphics[width=1.0\linewidth]{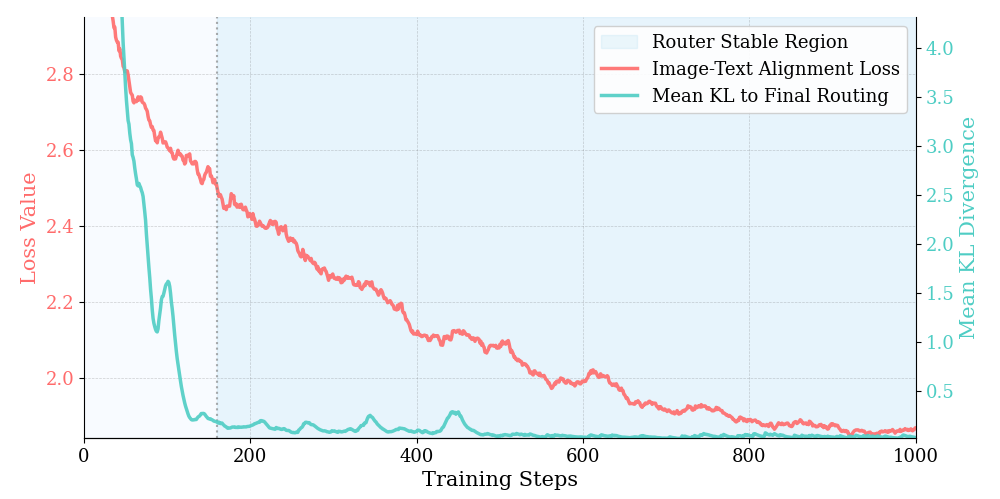}
   \caption{Training loss and mean KL divergence (averaged over multiple router layers) between the routing distributions at each snapshot and the final snapshot on the Cars dataset. The curves show that the multi-layer routing behavior converges much earlier than the image-text alignment loss. Additional results on more datasets are provided in the supplementary materials.}
   \label{fig:loss}
\end{figure}

Operationalizing this principle, we propose iGSP, a novel paradigm that formalizes continual learning as an implicit gradient subspace projection. iGSP dynamically navigates and shares cross-modal alignments using a Mixture-of-Experts (MoE) architecture. Crucially, we empirically observe that MoE routing distributions stabilize significantly earlier than the expert parameters during training (\cref{fig:loss}). From an optimization perspective, this early convergence indicates that the model rapidly identifies the optimal low-rank subspace basis required for the new task before fully minimizing the residual loss. Leveraging this fundamental property, iGSP naturally bifurcates the adaptation process into two tightly coupled phases separated by the router's convergence point: Subspace Discovery and Orthogonal Fine-tuning.
During the initial Subspace Discovery phase, we introduce a novel subspace-constrained regularization. Instead of explicitly computing computationally expensive Jacobian matrices, this regularization actively penalizes the utilization of new dimensions (new experts), implicitly forcing the optimizer to exhaust the expressive capacity of the existing expert subspace. New orthogonal dimensions are activated and updated only when the existing subspace yields a massive residual loss that overcomes the regularization penalty. Once the routing distribution converges, the basis of the subspace is effectively determined. By mathematically connecting routing frequency with gradient flow magnitude, iGSP prunes rarely activated candidates, systematically truncating redundant null-space dimensions in the gradient space.
Subsequently, in the Orthogonal Fine-tuning phase, since the optimal sharing structure is already fixed, the structural regularization is safely removed. The router is frozen, and the model exclusively updates the retained new experts. This allows the newly added orthogonal dimensions to rapidly fit the task-specific residual loss without structural interference or regularization drag. To handle realistic task-ID-free inference, iGSP incorporates ID-Free Expert Routing (IFER), which dynamically matches test representations to the appropriate optimization subspace.
The contributions of this paper are as follows:

\begin{itemize}
    \item We formalize cross-modal alignment reuse as an implicit geometric projection within gradient subspaces, exposing the fundamental flaws of heuristic visual-similarity-based sharing in continual learning.
    \item We propose iGSP, a two-stage framework that leverages early router convergence to automate subspace identification, redundancy truncation, and task-agnostic deployment, providing an end-to-end solution for efficient VLM adaptation.
    \item On the MTIL benchmark, iGSP achieves state-of-the-art accuracy while dramatically improving training efficiency, reducing the average number of trainable parameters by 42.7\% compared to current SOTA methods.
\end{itemize}
\section{Related Works}
\subsection{PTM-Based Continual Learning}
PTM-based continual learning\cite{ptmcontinual} adopts large pretrained backbones such as CLIP\cite{clip} and ViT\cite{vit} and learns over a stream of downstream tasks. Unlike general continual learning, PTM based methods must adapt while preserving the zero shot capability of the pretrained model\cite{zscl}. Classical approaches pursue knowledge compatibility: (i) Knowledge  distillation constrains the new model to match the old model on prior tasks \cite{zscl,lwf}; (ii) Regularization penalizes updates to parameters deemed important for previous tasks \cite{ewc} and (iii) Replay stores or synthesizes historical samples for rehearsal \cite{icarl,gift}. These strategies mitigate forgetting but often incur substantial compute or storage overhead, which limits practicality in compute constrained settings.

\subsection{Parameter-Efficient Fine-Tuning} 
In recent years, with the rapid expansion of large-scale pretrained models, full-parameter fine-tuning has achieved strong performance but suffers from high computational cost, large memory consumption, and deployment difficulties. To address these issues, Parameter-Efficient Fine-Tuning (PEFT) methods have been proposed\cite{prefix,lora}. 
To improve efficiency, recent work combines PEFT with continual learning by freezing the backbone and introducing small task specific parameter modules\cite{pecl1}, which reduces catastrophic forgetting. Methods fall into two families: prompt based\cite{l2p,dualprompt,s-prompts,feng2025pectp} and LoRA based\cite{inflora,cllora}. Prompt based methods encode task knowledge as learnable continuous vectors and focus on scalable prompt management. For example, L2P\cite{l2p} builds a shared prompt pool and uses key value retrieval to dynamically select and compose prompts per input; S-Prompts constructs task centroids via K means clustering and uses KNN to fetch the most relevant historical prompts for a new task \cite{s-prompts}. LoRA based methods attach independent low rank adaptation modules for each task; when learning a new task, only the current LoRA is trained while all historical modules are frozen to preserve prior knowledge. To further reduce interference, InfLoRA\cite{inflora} learns within a low rank subspace for the new task and enforces orthogonality to the gradient subspaces of past tasks . However, task specific adapters largely ignore the latent shared structure of cross task alignment. Although recent works \cite{sema,moeadapter} integrates MoE with adapters by treating adapters as experts and using routing to compose them at inference, these approaches still do not explicitly discover and exploit the shared structure to improve adaptation efficiency.

\subsection{Gradient Projection in Continual Learning} 
Gradient projection mitigates catastrophic forgetting from an optimization perspective by strictly confining the parameter updates of new tasks to the orthogonal null space of previous tasks. Early works such as GEM \cite{gem} and OGD \cite{ogd} laid the foundation by projecting gradients using explicit subspace bases. Recent advances have enriched this paradigm by integrating flatness-aware optimization \cite{yang2023data}, decoupling feature spaces into stability and plasticity sub-manifolds \cite{zhao2023rethinking}, or utilizing conceptor matrices \cite{apolinario2025code} as alternatives to standard feature covariance to construct more robust projection constraints.

As Parameter-Efficient Fine-Tuning (PEFT) becomes the standard for adapting large models, recent literature attempts to perform subspace optimization within restricted parameter spaces. In the continuous prompt space, methods like VPT-NS \cite{lu2024visual} and PGP \cite{qiao2024prompt} propose tuning visual prompts exclusively within the null space of prior tasks. For LoRA modules, KeepLoRA \cite{luo2026keeplora} restricts parameter updates to residual gradient subspaces to maintain backward stability, while SplitLoRA \cite{qiu2025splitlora} explicitly partitions the gradient space into orthogonal stability and plasticity components. 

In the context of Vision-Language Models (VLMs), explicitly preserving cross-modal alignment during continual adaptation is particularly critical. Recent state-of-the-art methods address this by strictly projecting task-specific gradients to avoid interference. For instance, GNSP \cite{peng2025gnsp} explicitly projects gradients onto the null spaces of past tasks to preserve alignment, and DMNSP \cite{kang2025dynamic} extends this via dynamic multi-layer null space projections. While conceptually related to our orthogonal subspace framing, these methods universally rely on strict structural isolation or computationally expensive explicit algebraic projections (e.g., Singular Value Decomposition on massive activation matrices). In contrast, our proposed iGSP formulates an \textit{implicit} gradient subspace projection. By substituting rigid algebraic decomposition with a subspace-constrained regularization (SCR) and leveraging the early convergence dynamics of MoE routers, iGSP acts as a "soft projection". It naturally achieves optimal stability-plasticity balance and discovers shared low-rank bases without SVD, substantially improving training efficiency while avoiding the parameter isolation problem.

\section{Methodology}
\label{sec:methods}
\begin{figure*}[htbp]
  \centering
   \includegraphics[width=1.0\linewidth]{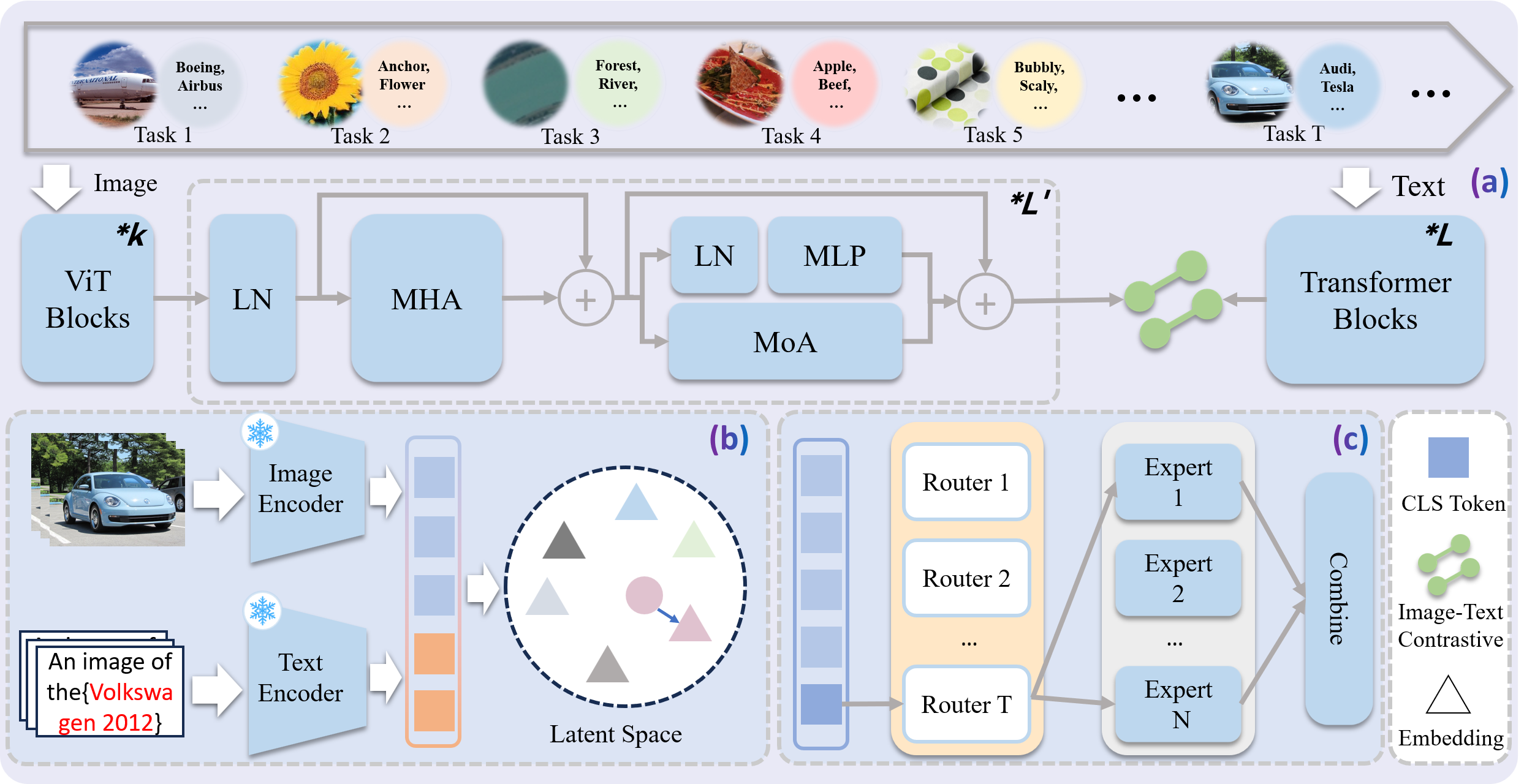}
   \caption{\textbf{(a)} Overall architecture of iGSP; \textbf{(b)} Detailed IFER pipeline; \textbf{(c)}MoE structure of the plug-in module.}
   \label{fig:framework}
\end{figure*}
\subsection{Problem Definition and Preliminaries}
\textbf{Continual Learning.}
Given a sequence of tasks $\mathcal{T} = \{T_1, T_2, T_3, \ldots, T_N\}$, each task $T_t = \{D_t, C_t\}$ consists of a dataset $D_t = \{(x_i^{(t)}, y_i^{(t)})\}_{i=1}^{n_t}$ and a category set $C_t = \{c_i^{(t)}\}_{i=1}^{m_t}$. Here, $x_i^{(t)}$ and $y_i^{(t)}$ denote the image and its corresponding label in task $T_t$, where labels are stored in one-hot encoding without inherent semantic meaning. $n_t$ and $m_t$ denote the total number of samples and categories in task $T_t$, respectively, while $C_t$ provides the semantic names of categories. For each task, the VLM learns the cross-modal alignment present in its training data\cite{aligment} by matching images to the natural-language names of the task’s categories. 

 During the training phase, tasks arrive sequentially. At time step $t$, the model can only access the current dataset $D_t$ and is prohibited from revisiting any previous datasets $\{D_1, \ldots, D_{t-1}\}$. During the testing phase, given an arbitrary sample $x$, the model must correctly predict its label $y$, assuming that the task boundary is unknown (\emph{i.e.}, task-ID free). The learning objective is to maximize the overall performance across all tasks after sequential training.

\textbf{Mixture of Experts.} A Mixture-of-Experts (MoE)\cite{moe} model is typically composed of a router $r$ and a set of experts $\{\varepsilon_j\}_{j=1}^{N_E}$. Given an input $x$, the router produces a routing distribution $\pi(x)$:
\begin{equation}
\pi(x) = [\pi_j(x)]_{j=1}^{N_E}, \quad \sum_j \pi_j(x) = 1
\end{equation}
where $\pi_j(x)$ denotes the activation probability of expert $\varepsilon_j$. The input $x$ is then forwarded to each expert network to obtain expert outputs $\varepsilon_j(x)$, where $j \in \{1, 2, \ldots, N_E\}$. The final output of the MoE is obtained by aggregating the expert outputs through a weighted average according to the routing probabilities:
\begin{equation}
y(x) = \sum_{j=1}^{N_E} \pi_j(x)\, \varepsilon_j(x).
\end{equation}

\textit{Optimization Perspective:} In our iGSP framework, each expert $\varepsilon_j$ is realized via a low-rank adapter (LoRA). Thus, the MoE layer acts as a dynamic integrator of parameter subspaces. Crucially, the routing probability $\pi_j(x)$ does not merely gate the forward pass but also scales the \textit{gradient flow magnitude} into each expert's parameter space during backpropagation. Consequently, the MoE structure can be viewed as a generator of task-specific gradient subspaces, where each expert serves as a potential \textit{basis vector} for cross-modal alignment.

\subsection{Framework Overview}

We propose a \textit{Implict Gradient Subspace Projection framework for continual learning} that achieves extensive expert sharing across tasks, substantially improving the parameter efficiency of existing continual learning methods. We argue that the adaptation of a new task should be treated as finding an optimal low-rank subspace within the gradient space. As illustrated in \cref{fig:framework}, iGSP is built upon the CLIP backbone and consists of a pretrained encoder followed by a multi-layer mixture adapter structure. Specifically, we insert mixture adapters into the last several layers of CLIP, from $h^{(k)}$ to $h^{(L)}$. Each mixture adapter comprises a set of routers and a set of expert networks, where each router is implemented by a linear layer followed by a Softmax function, and each expert adopts the LoRA-based adapter design.After training on $t$ tasks, the model can be represented as $M_t = \{ h_t^{(l)} \}_{l=k}^{L}$,where the $l$-th layer is defined as
\begin{equation}
    h_t^{(l)} = \{ \{ r_i^l \}_{i=1}^{N_R^{l,t}}, \{ \varepsilon_j^l \}_{j=1}^{N_E^{l,t}} \}.
\end{equation}
Here, $r_i^l$ and $\varepsilon_j^l$ denote the $i$-th router and the $j$-th expert network at layer $l$, respectively. $N_R^{l,t}$ and $N_E^{l,t}$ denote the numbers of routers and experts after $t$ training stages. The routers $r_i^l$ are task-specific; for a given task $T_t$, the $l$-th layer activates its corresponding router $r_t^l$ for inference.

\textbf{Training Phases.}
A critical property of MoE architectures is that the routing distribution stabilizes significantly earlier than the expert parameters (as empirically shown in \Cref{fig:loss}). From an optimization perspective, this early convergence signifies that the model rapidly identifies the required structural basis (the optimal subspace) before fully minimizing the residual loss. Leveraging this geometric property, iGSP naturally bifurcates the continual adaptation process into two tightly coupled stages: \textit{Subspace Identification} (Stage 1) and \textit{Orthogonal Subspace Fine-Tuning} (Stage 2).
The \textit{Subspace Identification} stage is executed through three systematic steps: (1) subspace basis pre-expansion, (2) rapid subspace identification, and (3) gradient-aware subspace truncation. This stage aims to discover the shared alignment by navigating the geometric span of previously learned tasks. To this end, we introduce \textit{Subspace-Constrained Regularization (SCR)}, which implicitly projects the gradient trajectory of the new task onto the historical experts. By penalizing the activation of new orthogonal dimensions, SCR forces the optimizer to maximize the reuse of established cross-modal alignments. 
In the subsequent \textit{Orthogonal Subspace Fine-Tuning} stage, the identified subspace basis is frozen, and the structural regularization is removed. This allows the model to concentrate exclusively on updating the newly retained experts to fit the task-specific residual loss. To prevent catastrophic forgetting, all previously learned experts remain frozen during both stages, serving as a stable geometric foundation while ensuring that new knowledge is accumulated only within independent orthogonal dimensions.

\textbf{Inference Phase.}
To enable task-ID-free inference, we further propose an \textit{ID-Free Expert Routing (IFER)} strategy. During inference, IFER consists of two key steps: (1) task identity estimation and (2) expert routing. This allows iGSP to perform accurate task-agnostic predictions without relying on explicit task identifiers.
\begin{figure*}[htbp]
  \centering
   \includegraphics[width=1.0\linewidth]{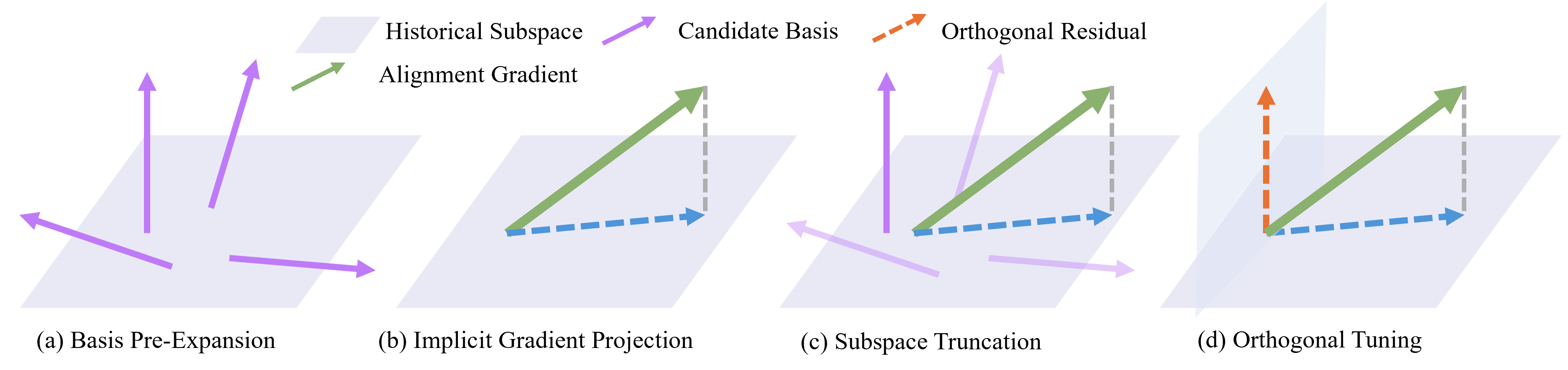}
   \caption{Geometric visualization of the iGSP optimization phases. (1) Subspace Pre-expansion: Initializing candidate basis vectors beyond the historical subspace. (2) Implicit Projection: SCR forces the task gradient to project onto the historical plane to maximize reuse. (3) Subspace Truncation: Redundant dimensions in the null-space are truncated based on gradient flow. (4) Orthogonal Fine-tuning: The minimal identified orthogonal basis is refined to fit the residual alignment..}
   \label{fig:sat}
\end{figure*}
\begin{figure*}[htbp]
  \centering
   \includegraphics[width=1.0\linewidth]{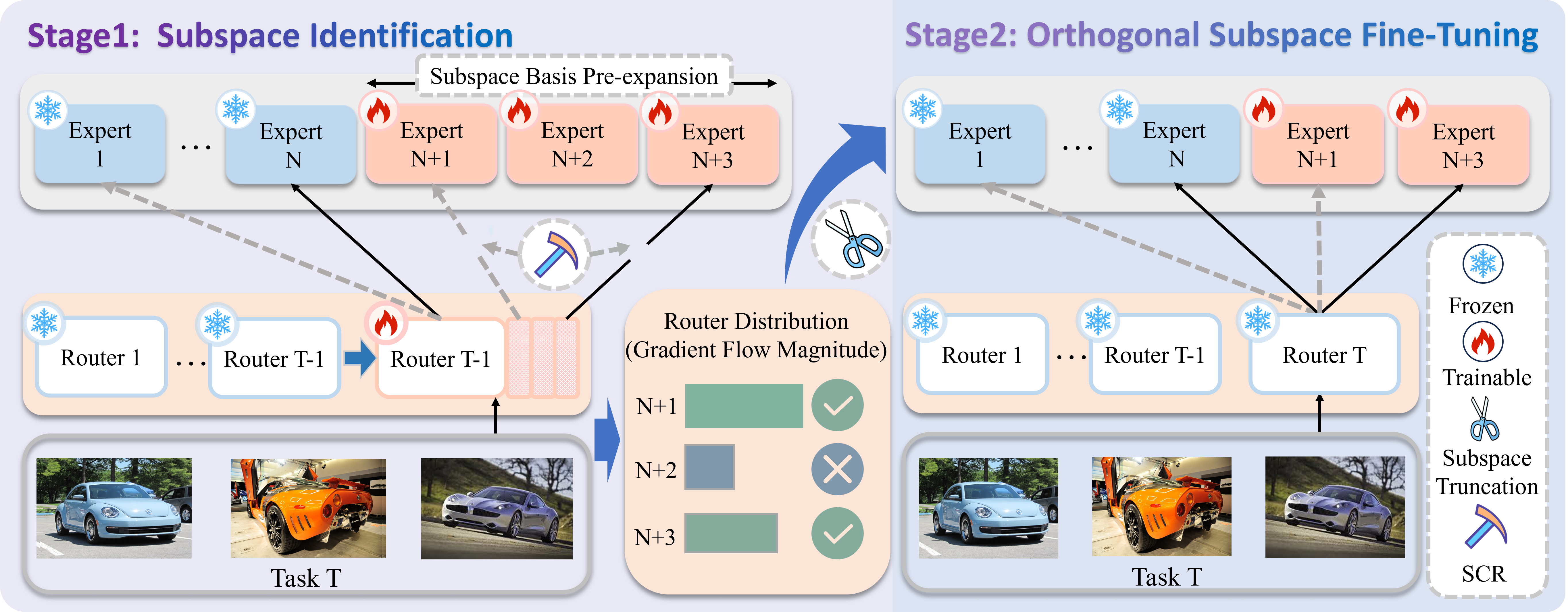}

   \caption{iGSP’s two-stage training procedure with the number of pre-expanded experts set to 3.}
   \label{fig:sat}
\end{figure*}
\subsection{Subspace Identification}
As illustrated in \cref{fig:sat}, the \textit{Subspace Identification} phase aims to determine the optimal low-rank structure for the new task by exploring the geometric span of available experts. iGSP adapts to the novel task by identifying a minimal set of basis vectors that maximizes the reuse of inter-task shared alignments, while strategically introducing a small number of candidate orthogonal dimensions to provide necessary representational headroom. 
The procedure initiates with \textit{subspace basis pre-expansion}, which initializes candidate experts as potential new dimensions. This is followed by \textit{rapid subspace identification} (driven by short-cycle training), where the router is encouraged to discover shared and essential new experts. Throughout this phase, we incorporate \textit{Subspace-Constrained Regularization (SCR)} to bias the gradient flow toward historical experts. By penalizing the activation of newly introduced dimensions, SCR implicitly forces the task to project its alignment logic onto the established cross-modal subspace, thereby fully exploiting prior knowledge. Finally, upon reaching the router's early convergence point, iGSP executes \textit{gradient-aware subspace truncation} to remove redundant candidate modules that reside in the gradient's null-space, resulting in a compact and efficient task-specific subspace.

\textbf{Subspace Basis Pre-expansion.}
When task $T_t$ arrives, the model has learned the previous $t\!-\!1$ tasks and is denoted by
$M^{t-1}=\{\,h_{t-1}^{(l)}\,\}_{l=k}^{L}$ where
$h_{t-1}^{(l)}=\big\{\, \{ \varepsilon_i^l \}_{i=1}^{N_E^{l,t-1}},\; \{ r_j^l \}_{j=1}^{N_R^{l,t-1}} \big\}$.
In the dynamic expansion stage, rather than blindly adding capacity, we first perform subspace basis pre-expansion. For each layer $h_{t-1}^{(l)}$, we add $M$ new expert modules and one additional router. Geometrically, these $M$ modules act as candidate orthogonal dimensions added to the existing subspace, yielding:
\begin{equation}
    h_{\mathrm{pre}}^{(l)}=\big\{\, \{ \varepsilon_i^l \}_{i=1}^{N_E^{l,t-1}+M},\; \{ r_j^l \}_{j=1}^{N_R^{l,t-1}+1} \big\}.
\end{equation}
Here, $M$ candidate basis vectors (experts) are introduced for $T_t$ to provide sufficient representational headroom for capturing the task-specific residual loss that cannot be projected onto the historical subspace. Notably, because the routing distribution becomes sparse after optimization, introducing multiple candidates does not materially inflate the number of active parameters. Concretely, at convergence, the routing probabilities 
$\pi^\ast(x^t)$ over these basis vectors are sparse: although $N_E$ candidates exist, the optimizer naturally concentrates the gradient flow onto a compact subset. This theoretically justifies our strategy of providing ample orthogonal dimensions upfront during pre-expansion, as the subsequent gradient-aware truncation will strictly prune the unutilized null-space dimensions.

\textbf{Rapid Subspace Identification.}
When optimizing the MoE on a new task $T_t$, the router distribution over the expanded candidate basis is highly volatile in early iterations. From an optimization perspective, this volatility indicates that the gradient trajectory is actively exploring the expanded high-dimensional space to identify the optimal parameter subspace. As updates accumulate, the routing decisions rapidly stabilize, converging to a task-specific low-rank subspace. Therefore, rather than fully minimizing the residual loss immediately, we execute a rapid subspace identification phase for $\Gamma$ steps. This brief exploration allows the optimization dynamics to reveal the intrinsic geometric structure of the task, yielding a converged routing distribution that dictates which orthogonal dimensions (pre-expanded experts) are strictly necessary to retain.

To better exploit inter-task sharing, we introduce a \textit{ Subspace-Constrained Regularization (SCR)} that biases the router toward shared experts and reduces reliance on newly added ones. This auxiliary loss penalizes the activation and update of new experts during training. For task $T_t$ at the $s$-th batch, the auxiliary loss at layer $l$ is
\begin{equation}
\mathcal{L}_{s}^{t,l}
= n \times \sum_{j = N_E^{t-1}+1}^{N_E^{t-1}+M}
\pi_{j}^{t,l}(x_s^t)
\big\| w_{s}^{(j)} - w_{s-1}^{(j)} \big\|_2^2,
\end{equation}
where $M$ is the number of new experts, $\pi_{j}^{t,l}(x_s^t)$ is the routing probability at layer $l$ for expert $\varepsilon_j^l$ on input $x_s^t$, and $w_{s}^{(\varepsilon_j^l)}$ and $w_{s-1}^{(\varepsilon_j^l)}$ are the parameters of $\varepsilon_j^l$ at steps $s$ and $s\!-\!1$, respectively; $n$ denotes the number of experts whose weights change between steps $s\!-\!1$ and $s$. The auxiliary loss on task $T_t$ is computed on the visual encoder and aggregated across its layers, denoted as $\mathcal{L}^{\mathrm{aux},t}$. The overall training objective is
\begin{equation}
\mathcal{L}
= \mathcal{L}_{\mathrm{Contrastive}}
+ \lambda\,\mathcal{L}^{\mathrm{aux},t},
\end{equation}
where $\mathcal{L}_{\mathrm{Contrastive}}$ is the standard CLIP contrastive (cross-entropy) loss, and $\lambda$ controls the strength of the auxiliary regularization.

\textbf{Theoretical Perspective: SCR as Implicit Subspace Projection.} 
Mathematically, we can rigorously demonstrate that the proposed Subspace-Constrained Regularization (SCR) operates as an \textit{implicit gradient projection} via the lens of proximal gradient descent. We conceptually partition the global parameter space of the MoE layer into two orthogonal subspaces: the shared subspace $\mathcal{S}_{\mathrm{old}}$ spanned by the existing experts, and the expanded exploration subspace $\mathcal{S}_{\mathrm{new}}$ spanned by the newly added experts. It is imperative to emphasize that this orthogonality is strictly defined within the Euclidean parameter space, rather than the output feature space. Specifically, by flattening the parameters into a high-dimensional vector, the total parameter space $\mathcal{W}$ can be formulated as the direct sum $\mathcal{W} = W_{\mathrm{old}} \oplus W_{\mathrm{new}}$. From a block-coordinate optimization perspective, any parameter state in $\mathcal{S}_{\mathrm{old}}$ and $\mathcal{S}_{\mathrm{new}}$ takes the block-vector form of $[W_{\mathrm{old}}^\top, \mathbf{0}^\top]^\top$ and $[\mathbf{0}^\top, W_{\mathrm{new}}^\top]^\top$, respectively. Their Euclidean inner product is intrinsically zero:
\begin{equation}
    \big\langle[W_{\mathrm{old}}^\top, \mathbf{0}^\top]^\top,[\mathbf{0}^\top, W_{\mathrm{new}}^\top]^\top \big\rangle = W_{\mathrm{old}}^\top \mathbf{0} + \mathbf{0}^\top W_{\mathrm{new}} \equiv 0,
\end{equation}
rendering the two subspaces naturally and mutually orthogonal. Let $g_s = \nabla \mathcal{L}_{\mathrm{Contrastive}}(w_{s-1})$ denote the raw gradient of the contrastive loss. When employing standard Stochastic Gradient Descent (SGD) with a learning rate $\eta$, the inclusion of the auxiliary penalty $\mathcal{L}^{\mathrm{aux},t}$ effectively transforms the parameter update for a new expert $j \in \mathcal{S}_{\mathrm{new}}$ into solving the following localized proximal minimization problem at step $s$:
\begin{equation}
\begin{split}
w_s^{(j)} &= \arg\min_{w} \bigg[ g_s^{(j)\top} \big(w - w_{s-1}^{(j)}\big) \\
&\quad + \frac{1}{2\eta} \big\| w - w_{s-1}^{(j)} \big\|_2^2 + \lambda n \pi_{j}^{t,l}(x_s^t) \big\| w - w_{s-1}^{(j)} \big\|_2^2 \bigg].
\end{split}
\end{equation}
Taking the derivative with respect to $w$ and setting it to zero yields:
\begin{equation}
    g_s^{(j)} + \frac{1}{\eta} \big(w_s^{(j)} - w_{s-1}^{(j)}\big) + 2 \lambda n \pi_{j}^{t,l}(x_s^t) \big(w_s^{(j)} - w_{s-1}^{(j)}\big) = 0.
\end{equation}
By solving for the actual parameter update $\Delta w^{(j)} = w_s^{(j)} - w_{s-1}^{(j)}$, we obtain:
\begin{equation}
    \Delta w^{(j)} = - \frac{\eta}{1 + 2\eta \lambda n \pi_{j}^{t,l}(x_s^t)} g_s^{(j)}.
\end{equation}
Concurrently, since the old experts $i \in \mathcal{S}_{\mathrm{old}}$ are excluded from the SCR penalty, their updates remain unconstrained, i.e., $\Delta w^{(i)} = -\eta g_s^{(i)}$. By concatenating the gradients as $g_s =[g_{\mathrm{old}}^\top, g_{\mathrm{new}}^\top]^\top$, the global update dynamic can be elegantly formulated as a matrix-vector product:
\begin{equation}
\begin{split}
\Delta W &= - \eta P_{\mathrm{implicit}} g_s, \\
\text{where } P_{\mathrm{implicit}} &= \begin{pmatrix} \mathbf{I}_{\mathrm{old}} & \mathbf{0} \\ \mathbf{0} & \mathbf{\Gamma}_{\mathrm{new}} \end{pmatrix}.
\end{split}
\end{equation}
with $\mathbf{\Gamma}_{\mathrm{new}} = \mathrm{diag}\left( \frac{1}{1 + 2\eta \lambda n \pi_{j}^{t,l}(x_s^t)} \right)_{j=N_E^{t-1}+1}^{N_E^{t-1}+M}$. 

Crucially, an ideal strict orthogonal projection onto the old experts' subspace $\mathcal{S}_{\mathrm{old}}$ would demand a projection matrix $P_{\mathrm{strict}} = \mathrm{diag}(\mathbf{I}_{\mathrm{old}}, \mathbf{0}_{\mathrm{new}})$. Our derived $P_{\mathrm{implicit}}$ reveals a data-dependent, soft projection mechanism. Whenever the router actively distributes probability mass to a new expert (i.e., $\pi_{j}^{t,l} \gg 0$), the accumulated scaling term $2\eta \lambda n \pi_{j}^{t,l}$ becomes non-negligible. This forces the corresponding diagonal entry in $\mathbf{\Gamma}_{\mathrm{new}}$ to be strictly bounded below $1$, effectively dampening the update step. In the asymptotic limit, this dynamic approaches a strict projection $P_{\mathrm{implicit}} \approx P_{\mathrm{strict}}$. Unlike explicit hard-projection algorithms that require computationally expensive Singular Value Decomposition (SVD), our SCR gracefully acts as an anisotropic shrinkage operator. It seamlessly suppresses the divergent gradient flows in the high-dimensional $\mathcal{S}_{\mathrm{new}}$ and constrains the optimization trajectory tightly within the low-rank shared subspace $\mathcal{S}_{\mathrm{old}}$, thereby yielding a mathematically sound realization of the aforementioned rapid subspace identification.

\textbf{Gradient-Aware Subspace Truncation.}
Once the routing distribution converges, the basis of the task-specific subspace is structurally determined. At this point, iGSP executes \textit{Gradient-Aware Subspace Truncation} to remove unnecessary candidate modules. Crucially, the pruning is not a heuristic guess but is mathematically grounded in gradient dynamics. During backpropagation, the gradient of the loss $\mathcal{L}$ with respect to an expert $E_j$'s parameters $\theta_j$ is determined by the chain rule:
\begin{equation}
\nabla_{\theta_j} \mathcal{L} = \frac{\partial \mathcal{L}}{\partial y} \cdot \pi_j(x) \cdot \frac{\partial E_j(x)}{\partial \theta_j},
\end{equation}
where $y$ denotes the output of the mixture-of-experts layer. This formulation reveals a fundamental property: the routing probability $\pi_j(x)$ directly mirrors the \textit{gradient flow magnitude} along the direction of expert $j$. If the converged routing probability $\pi_j^*$ for a candidate expert falls below a threshold $\tau$, it mathematically indicates that the gradient flow in this orthogonal dimension approaches zero. Therefore, this expert represents a \textit{redundant null-space dimension} for the current task. By removing experts where $\pi_j^* < \tau$, iGSP systematically truncates these null-space dimensions, drastically reducing parameter redundancy without sacrificing representational capacity.If $X$ experts are removed, the pruned layer becomes
\begin{equation}
h_{t}^{(l)}=\big\{\, \{ \varepsilon_i^l \}_{i=1}^{N_E^{l,t-1}+M-X},\; \{ r_j^l \}_{j=1}^{N_R^{l,t-1}+1} \big\}.
\end{equation}

Finally, for task $T_t$, iGSP yields the expanded-and-pruned model
$M^{t}=\{\,h_{t}^{(l)}\,\}_{l=k}^{L}$.

Mathematically, this truncation process transforms the over-complete candidate basis set into a minimal basis set that spans the optimal task-specific subspace, effectively filtering out dimensions with near-zero gradient energy.

\subsection{Orthogonal Subspace Fine-Tuning}
Following the truncation of null-space dimensions, the optimal sharing structure and the required new basis vectors are finalized. iGSP then transitions to the Orthogonal Subspace Fine-Tuning phase.
Because the structural search is complete, the Subspace-Constrained Regularization (SCR) is safely removed. We freeze the router (fixing the subspace basis) and exclusively update the weights of the retained new experts. Geometric Motivation: Removing the regularization in this phase is crucial. Since the newly retained experts have been identified as necessary orthogonal dimensions to satisfy the task-specific residual loss, continuing to penalize them would act as an optimization drag. By isolating these dimensions and allowing unconstrained gradient descent, the newly added experts rapidly and accurately fit the task-specific data distribution without causing catastrophic interference to the frozen historical subspace.

\subsection{ID-Free Expert Routing (IFER)}

Because each router is bound to a specific task (i.e., router $r_i$ is used when testing task $T_i$), direct inference does not meet real-world requirements where task IDs are unavailable. We therefore propose \textit{ID-Free Expert Routing (IFER)}, which first \emph{infers the task identity} of the test input and then either activates the corresponding router or falls back to the frozen pretrained backbone for prediction.

\textbf{Task Identity Inference.}
IFER uses the frozen image--text encoders $(E_V,E_L)$ to embed images and labels, introducing no extra trainable parameters. For each task $T_t$, we sample a mini-batch of images and all candidate labels, obtain embeddings $f_{\mathrm{img}}^t$ and $f_{\mathrm{txt}}^t$, and form a fused task embedding by concatenating their mean-pooled features:
\begin{equation}
    f^t = \mathrm{Concat}\big(\mathrm{Mean}(f_{\mathrm{img}}^t),\, \mathrm{Mean}(f_{\mathrm{txt}}^t)\big).
\end{equation}
All $\{f^t\}$ are stored in a task embedding bank. At test time, we build a query embedding in the same way and retrieve the nearest neighbor in the bank; if the minimum distance exceeds a threshold $\delta$, the query is regarded as unmatched.

\textbf{Expert Routing.}
If the query is matched to task $T_{i^\ast}$, iGSP activates the corresponding routers $\{r_{i^\ast}^l\}$, which perform top-$k$ gating over experts at each layer. If the query is unmatched, iGSP falls back to the frozen backbone only, bypassing all task-specific routers.

\section{Experiments }
\subsection{Experimental Setup}
\textbf{Datasets and Benchmarks.}
To evaluate the efficacy of the proposed iGSP, we conduct extensive experiments across various incremental learning scenarios and benchmarks. We utilize three primary benchmarks to assess performance under different settings:
\begin{enumerate}
    \item MTIL and MTIL-FS: The Multi-Task Incremental Learning (MTIL) benchmark \cite{zscl} consists of 11 diverse datasets: Aircraft, Caltech101, CIFAR100, DTD, EuroSAT, Flowers, Food, MNIST, OxfordPet, StanfordCars, and SUN397. These datasets cover a wide range of domains, from natural objects to satellite imagery. 
    \item CIL Benchmarks: For the Class-Incremental Learning (CIL) scenario, we employ two widely-used datasets: \textbf{CIFAR-100} \cite{cifar100} and \textbf{TinyImageNet} \cite{der}. CIFAR-100 contains 100 classes, while TinyImageNet consists of 200 classes with higher-resolution images.
\end{enumerate}

\textbf{Task Settings.}
We evaluate our model under both task-incremental and class-incremental settings:
\begin{enumerate}
 \item Task-Incremental Scenario (MTIL): Following \cite{zscl}, we organize the 11 datasets into two distinct sequences to introduce different domain shifts. The first sequence, referred to as Order I, follows alphabetical order:Aircraft , Caltech101 , CIFAR100 , DTD , EuroSAT , Flowers , Food , MNIST , OxfordPet , StanfordCars , SUN397. The second sequence, Order II, is randomly arranged:StanfordCars , Food , MNIST , OxfordPet , Flowers , SUN397 , Aircraft , Caltech101 , DTD , EuroSAT , CIFAR100.

 \item Class-Incremental Scenario (CIL): To test the scalability of our approach, CIFAR-100 is partitioned into 10, 20, and 50 disjoint subsets (tasks). Similarly, TinyImageNet is partitioned into 5, 10, and 20 subsets. In this setting, the model must classify all classes seen so far without knowing the task identity during inference.
\end{enumerate}

\textbf{Evaluation Metrics.}
For the MTIL, let $a_{i,j}$ denote the test accuracy on task $j$ after the model has learned task $i$, where $i, j \in \{1, \dots, n\}$ and $n$ is the total number of tasks. In the context of Vision-Language Models (VLMs), we utilize the full accuracy matrix $[a_{i,j}]_{n \times n}$ to compute three key metrics:

\begin{enumerate}
    \item \textbf{Transfer:} Measures the model's zero-shot transferability to unseen tasks (the upper triangular part of the matrix):
    \begin{equation}
        \text{Transfer} = \frac{1}{n-1} \sum_{j=2}^{n} \frac{1}{j-1} \sum_{i=1}^{j-1} a_{i,j}
    \end{equation}
    
    \item \textbf{Last:} Evaluates the final performance and the ability to retain knowledge of all learned tasks after the incremental process:
    \begin{equation}
        \text{Last} = \frac{1}{n} \sum_{j=1}^{n} a_{n,j}
    \end{equation}
    
    \item \textbf{Average (Avg.):} Provides a holistic measure of performance throughout the learning process, considering both learned and unlearned tasks:
    \begin{equation}
        \text{Avg.} = \frac{1}{n^2} \sum_{i=1}^{n} \sum_{j=1}^{n} a_{i,j}
    \end{equation}
\end{enumerate}

For the CIL setting, the model is evaluated on the test set containing all classes observed so far after each incremental step. Let $A_i$ denote the classification accuracy after learning the $i$-th task, evaluated on the test set containing all classes from tasks $1$ to $i$.

\begin{enumerate}
    \item \textbf{Last:} Measures the final classification accuracy after the model has learned all tasks:
    \begin{equation}
        \text{Last}= A_n
    \end{equation}

    \item \textbf{Average (Avg.):} Measures the average incremental accuracy across all learning stages:
    \begin{equation}
        \text{Avg.} = \frac{1}{n} \sum_{i=1}^{n} A_i
    \end{equation}
\end{enumerate} 

\textbf{Implementation Details.} We build iGSP on top of CLIP and adopt ViT-B/16 as the visual backbone. 
We use AdamW\cite{adamW} as the optimizer with a learning rate of $10^{-2}$ for all tasks. 
The MoE module adopts top-2 routing, with the number of pre-expanded experts set to 1 and an expert pruning threshold of $\tau = 0.1$. 
For full-shot tasks, we train for 1000 epochs in total, including 500 epochs for the expert combination search phase and 500 epochs for expert fine-tuning. 
For few-shot tasks, we train for 500 epochs in total, with 200 epochs for expert combination search and 300 epochs for the final training stage. 
In the IFER module, we use the Manhattan distance as the similarity measure and set the task-identification threshold to 10.

\begin{table}[htbp]
\centering
\setlength{\tabcolsep}{4pt}
\footnotesize
\caption{Comparison of SOTA methods on the MTIL Order-I.}
\begin{tabular}{lcccccc}
\toprule
\multirow{2}{*}{Method} &
\multicolumn{2}{c}{Transfer $\Delta$} &
\multicolumn{2}{c}{Avg. $\Delta$} &
\multicolumn{2}{c}{Last $\Delta$} \\
\cmidrule(lr){2-3} \cmidrule(lr){4-5} \cmidrule(lr){6-7}
 & Value & $\Delta$ & Value & $\Delta$ & Value & $\Delta$ \\
\midrule
\quad Zero-shot & 69.4 & 0.0 & 65.3 & 0.0 & 65.3 & 0.0 \\
\quad Continual-FT & 44.6 & {\color{blue}-24.8} & 55.9 & {\color{blue}-9.4} & 77.3 & {\color{red}+12.0} \\
\quad LwF~\cite{lwf} & 58.9 & {\color{blue}-10.5} & 64.7 & {\color{blue}-0.6} & 74.6 & {\color{red}+9.3} \\
\quad iCaRL~\cite{icarl} & 50.4 & {\color{blue}-19.0} & 65.7 & {\color{red}+0.4} & 80.1 & {\color{red}+14.8} \\
\quad LwF-VR~\cite{lwf-vr} & 57.2 & {\color{blue}-12.2} & 65.1 & {\color{blue}-0.2} & 76.6 & {\color{red}+11.3} \\
\quad WiSE-FT~\cite{wise-ft} & 52.3 & {\color{blue}-17.1} & 60.7 & {\color{blue}-4.6} & 77.7 & {\color{red}+12.4} \\
\quad ZSCL~\cite{zscl} & 68.1 & {\color{blue}-1.3} & 75.4 & {\color{red}+10.1} & 83.6 & {\color{red}+18.3} \\
\quad L2P~\cite{l2p} & 53.2 & {\color{blue}-16.2} & 67.9 & {\color{red}+2.6} & 82.0 & {\color{red}+16.7} \\
\quad Dual-prompt~\cite{dualprompt} & 52.4 & {\color{blue}-17.0} & 68.0 & {\color{red}+2.7} & 82.3 & {\color{red}+17.0} \\
\quad S-Prompts~\cite{s-prompts} & 52.2 & {\color{blue}-17.2} & 68.3 & {\color{red}+3.0} & 83.4 & {\color{red}+18.1} \\
\quad DIKI~\cite{diki} & 68.7 & {\color{blue}-0.7} & 76.3 & {\color{red}+11.0} & 85.1 & {\color{red}+19.8} \\
\quad MoE-Adapter~\cite{moeadapter} & 68.9 & {\color{blue}-0.5} & 76.7 & {\color{red}+11.4} & 85.0 & {\color{red}+19.7} \\
\quad MoE-Adapter++~\cite{yu2025moe} & \underline{69.0} & \underline{{\color{blue}-0.4}} & \underline{77.5} & \underline{{\color{red}+12.2}} & \underline{86.2} & \underline{{\color{red}+20.9}} \\
\midrule
\rowcolor[RGB]{220,228,244}
\quad \textbf{iGSP (Ours)} &
\textbf{69.9} & {\color{red}\textbf{+0.5}} &
\textbf{78.0}& {\color{red}\textbf{+12.7}}&
\textbf{86.7}& {\color{red}\textbf{+21.4}}\\
\bottomrule
\end{tabular}
\label{tab:mtilo1}
\end{table}
\subsection{Comparison with State-of-the-art Methods}
\textbf {Results on Multi-domain Task Incremental Learning.}
\cref{tab:mtilo1} and \cref{tab:mtilo2} compare our proposed iGSP with several state-of-the-art (SOTA) continual learning methods under different MTIL configurations, including both Order-I and Order-II. 
All methods are evaluated using three metrics: \textit{Transfer}, \textit{Average}, and \textit{Last}. 
The best results for each metric are highlighted in \textbf{bold}, and the second-best ones are \underline{underlined}. 
\textit{Continual-FT} denotes continual fine-tuning without any forgetting mitigation mechanism.

From the boldfaced results, iGSP consistently achieves the best performance across all metrics and task orders. 
Under the Order-I configuration, iGSP surpasses the previous SOTA by 0.9\%, 0.5\%, and 0.5\% in \textit{Transfer}, \textit{Average} and \textit{Last}, respectively. 
Under the Order-II configuration, it further improves upon the best competitor by 1\%, 1.4\% and 1.5\%  in \textit{Transfer} \textit{Average} and \textit{Last}, demonstrating robust performance even when the task sequence is rearranged. 
\begin{table}[htbp]
\centering
\setlength{\tabcolsep}{4pt}
\footnotesize
\caption{Comparison of SOTA methods on the MTIL Order-II.}
\begin{tabular}{lcccccc}
\toprule
\multirow{2}{*}{Method} &
\multicolumn{2}{c}{Transfer $\Delta$} &
\multicolumn{2}{c}{Avg. $\Delta$} &
\multicolumn{2}{c}{Last $\Delta$} \\
\cmidrule(lr){2-3} \cmidrule(lr){4-5} \cmidrule(lr){6-7}
 & Value & $\Delta$ & Value & $\Delta$ & Value & $\Delta$ \\
\midrule
\quad Zero-shot & 65.4 & 0.0 & 65.3 & 0.0 & 65.3 & 0.0 \\
\quad Continual-FT& 46.6 & {\color{blue}-18.8} & 56.2 & {\color{blue}-9.1} & 67.4 & {\color{red}+2.1} \\
\quad LwF~\cite{lwf} & 53.2 & {\color{blue}-12.2} & 62.2 & {\color{blue}-3.1} & 71.9 & {\color{red}+6.6} \\
\quad iCaRL~\cite{icarl} & 50.9 & {\color{blue}-14.5} & 56.9 & {\color{blue}-8.4} & 71.6 & {\color{red}+6.3} \\
\quad LwF-VR~\cite{lwf-vr} & 53.1 & {\color{blue}-12.3} & 60.6 & {\color{blue}-4.7} & 68.2 & {\color{red}+2.9} \\
\quad WiSE-FT~\cite{wise-ft} & 51.1 & {\color{blue}-14.3} & 61.5 & {\color{blue}-3.8} & 72.2 & {\color{red}+6.9} \\
\quad ZSCL~\cite{zscl} & 64.1 & {\color{blue}-1.3} & 74.5 & {\color{red}+9.2} & 83.4 & {\color{red}+18.1} \\
\quad L2P~\cite{l2p} & 42.5 & {\color{blue}-22.9} & 62.5 & {\color{blue}-2.8} & 82.3 & {\color{red}+17.0} \\
\quad DualPrompt~\cite{dualprompt} & 52.1 & {\color{blue}-13.3} & 67.5 & {\color{red}+2.2} & 82.8 & {\color{red}+17.5} \\
\quad S-Prompts~\cite{s-prompts} & 45.3 & {\color{blue}-20.1} & 65.1 & {\color{blue}-0.2} & 83.8 & {\color{red}+18.5} \\
\quad DIKI~\cite{diki} & 64.4 & {\color{blue}-1.0} & 74.5 & {\color{red}+9.2} & 85.5 & {\color{red}+20.2} \\
\quad MoE-Adapter~\cite{moeadapter} & 64.3 & {\color{blue}-1.1} & 74.7 & {\color{red}+9.4} & 84.1 & {\color{red}+18.8} \\
\quad MoE-Adapter++~\cite{yu2025moe} & \underline{64.8} & \underline{{\color{blue}-0.6}} & \underline{75.1} & \underline{{\color{red}+9.8}} & \underline{85.6} & \underline{{\color{red}+20.3}} \\
\midrule
\rowcolor[RGB]{220,228,244}
\quad \textbf{iGSP (Ours)} &
\textbf{65.8} & {\color{red}\textbf{+0.4}} &
\textbf{76.5}& {\color{red}\textbf{+11.2}}&
\textbf{87.1}& {\color{red}\textbf{+21.8}}\\
\bottomrule
\end{tabular}
\label{tab:mtilo2}
\end{table}

\textbf {Results on Few Shot Multi-domain Task Incremental Learning.}
As shown in \cref{tab:mtil-fso1} and \cref{tab:mtil-fso2}, iGSP also achieves the highest overall scores under the few-shot setting.
In the Order-I configuration, iGSP improves the previous SOTA by 0.6\% and 0.2\% on \textit{Transfer} and \textit{Average}, respectively, while remaining comparable on \textit{Last} with a marginal 0.2\% difference.
In the Order-II configuration, iGSP further outperforms the prior best by 0.6\%, 0.4\%, and 0.3\% on \textit{Transfer}, \textit{Average}, and \textit{Last}, respectively.
These consistent improvements across both task orders and data regimes indicate that iGSP maintains superior stability and adaptability, highlighting its robustness to task-sequence variations. 
\begin{table}[htbp]
\centering
\caption{Comparison of SOTA methods on the MTIL-FS-OrderI.}
\setlength{\tabcolsep}{4pt}
\footnotesize
\centering
\begin{tabular}{lcccccc}
\toprule
\multirow{2}{*}{Method} & \multicolumn{2}{c}{Transfer $\Delta$} & \multicolumn{2}{c}{Avg. $\Delta$} & \multicolumn{2}{c}{Last $\Delta$} \\
\cmidrule(lr){2-3} \cmidrule(lr){4-5} \cmidrule(lr){6-7}
 & Value & $\Delta$ & Value & $\Delta$ & Value & $\Delta$ \\
\midrule
\quad Zero-shot & 69.4 & 0.0 & 65.3 & 0.0 & 65.3 & 0.0 \\
\quad Continual-FT & 51.2 & {\color{blue}-18.2} & 58.5 & {\color{blue}-6.8} & 67.1 & {\color{red}+1.8} \\
\quad LwF~\cite{lwf} & 50.0 & {\color{blue}-19.4} & 49.2 & {\color{blue}-16.1} & 46.5 & {\color{blue}-18.8} \\
\quad LwF-VR~\cite{lwf-vr} & 60.1 & {\color{blue}-9.3} & 62.5 & {\color{blue}-2.8} & 66.9 & {\color{blue}-0.4} \\
\quad WiSE-FT~\cite{wise-ft} & 57.7 & {\color{blue}-11.7} & 63.7 & {\color{blue}-1.6} & 71.9 & {\color{red}+6.6} \\
\quad ZSCL~\cite{zscl} & 65.3 & {\color{blue}-4.1} & 64.4 & {\color{blue}-0.9} & 67.4 & {\color{red}+2.1} \\
\quad MoE-Adapter~\cite{moeadapter} & 68.9 & {\color{blue}-0.5} & 71.4 & {\color{red}+6.1} & 76.1 & {\color{red}+10.8} \\
\quad MoE-Adapter++~\cite{yu2025moe} &\underline{69.3} & \underline{{\color{blue}-0.1}} & \underline{71.7} & \underline{{\color{red}+6.4}} &\textbf{76.3} &\textbf{{\color{red}+11.0}} \\
\midrule
\rowcolor[RGB]{220,228,244}
\quad \textbf{iGSP (Ours)} & \textbf{69.9} & {\color{red}\textbf{+0.5}} & \textbf{71.9} & {\color{red}\textbf{+6.6}} & \underline{76.1} & {\color{red}\underline{+10.8}} \\
\bottomrule
\end{tabular}
\label{tab:mtil-fso1}
\end{table}
\begin{table}[htbp]
\caption{Comparison of SOTA methods on the MTIL-FS-OrderII.}
\setlength{\tabcolsep}{4pt}
\centering
\begin{tabular}{lcccccc}
\toprule
\multirow{2}{*}{Method} &
\multicolumn{2}{c}{Transfer $\Delta$} &
\multicolumn{2}{c}{Avg. $\Delta$} &
\multicolumn{2}{c}{Last $\Delta$} \\
\cmidrule(lr){2-3} \cmidrule(lr){4-5} \cmidrule(lr){6-7}
 & Value & $\Delta$ & Value & $\Delta$ & Value & $\Delta$ \\
\midrule
\quad Zero-shot & 65.4& 0.0 & 65.3 & 0.0 & 65.3 & 0.0 \\
\quad Continual-FT & 49.2 & {\color{blue}-16.2}& 49.0 & {\color{blue}-16.3} & 42.8 & {\color{blue}-22.5} \\
\quad LwF~\cite{lwf} & 48.6 & {\color{blue}-16.8}& 54.7 & {\color{blue}-10.6} & 56.8 & {\color{blue}-8.5} \\
\quad LwF-VR~\cite{lwf-vr} & 54.4 & {\color{blue}-11.0}& 60.1 & {\color{blue}-5.2} & 63.5 & {\color{blue}-1.8} \\
\quad WiSE-FT~\cite{wise-ft} & 53.7 & {\color{blue}-11.7}& 55.1 & {\color{blue}-10.2} & 51.8 & {\color{blue}-13.5} \\
\quad ZSCL~\cite{zscl} & 62.8 & {\color{blue}-2.6}& 67.6 & {\color{red}+2.3} & 71.8 & {\color{red}+6.5} \\
\quad MoE-Adapter~\cite{moeadapter} & 64.7 & {\color{blue}-0.7}& 69.5 & {\color{red}+4.2} & 75.7 & {\color{red}+10.4} \\
\quad MoE-Adapter++~\cite{yu2025moe} &\underline{65.2} & \underline{{\color{blue}-0.2}} & \underline{70.4} & \underline{{\color{red}+5.1}} &\underline{76.3} &\underline{{\color{red}+11.0}} \\
\midrule
\rowcolor[RGB]{220,228,244}
\quad \textbf{iGSP (Ours)} &
\textbf{65.8}& {\color{red}\textbf{+0.4}}&
\textbf{70.8}& {\color{red}\textbf{+5.5}}&
\textbf{76.6}& {\color{red}\textbf{+11.3}}\\
\bottomrule
\end{tabular}
\label{tab:mtil-fso2}
\end{table}

\textbf{Results on CIL Benchmarks}
We conduct experiments in the class incremental learning (CIL) setting to evaluate the proposed method on single-domain continual learning. Unlike MTIL, the task ID of the input image is unknown in CL. Following the design of MoE-Adapters, we employ a single router with two experts to adapt to all subsets. We compare our approach with state-of-the-art methods on TinyImageNet and CIFAR100, with the corresponding results reported in \cref{tab:tinyimage} and \cref{tab:cifar}, respectively. As can be seen, the proposed method achieves the best performance in the vast majority of settings.

\begin{table}[htbp]
\caption{Comparison of different methods on TinyImageNet splits in class-incremental settings with 100 base classes.}
\centering
\footnotesize
\setlength{\tabcolsep}{4pt}

\begin{tabular}{lcccccc}
\toprule
\multirow{2}{*}{Method} &
\multicolumn{2}{c}{5-step} &
\multicolumn{2}{c}{10-step} &
\multicolumn{2}{c}{20-step} \\
\cmidrule(lr){2-3} \cmidrule(lr){4-5} \cmidrule(lr){6-7}
& Avg. & Last & Avg. & Last & Avg. & Last \\
\midrule
EWC\cite{ewc}        & 19.01 &  6.00 & 15.82 &  3.79 & 12.35 &  4.73 \\
EEIL\cite{eeil}        & 47.17 & 35.12 & 45.03 & 34.64 & 40.41 & 29.72 \\
UCIR\cite{ucir}       & 50.30 & 39.42 & 48.58 & 37.29 & 42.84 & 30.85 \\
MUC\cite{muc}        & 32.23 & 19.20 & 26.67 & 15.33 & 21.89 & 10.32 \\
PASS\cite{pass}       & 49.54 & 41.64 & 47.19 & 39.27 & 42.01 & 32.93 \\
DyTox+\cite{dytox}      & 55.58 & 47.23 & 52.26 & 42.79 & 46.18 & 36.21 \\
\midrule
CLIP zero-shot  & 69.62 & 65.30 & 69.55 & 65.59 & 69.49 & 65.30 \\
Fine-tune       & 61.54 & 46.66 & 57.05 & 41.54 & 54.62 & 44.55 \\
LwF\cite{lwf}        & 60.97 & 48.77 & 57.60 & 44.00 & 54.79 & 42.26 \\
iCaRL\cite{icarl}      & 77.02 & 70.39 & 73.48 & 65.97 & 69.65 & 64.68 \\
LwF-VR\cite{lwf-vr}     & 77.56 & 70.89 & 74.12 & 67.05 & 69.94 & 63.89 \\
ZSCL\cite{zscl}       & 80.27 & 73.57 & 78.61 & \underline{71.62} & 77.18 & 68.30 \\
MoE-Adapter(++)\cite{moeadapter} &\underline{81.12} &\underline{76.81} &\underline{80.23}&\textbf{76.35} &\underline{79.96} &\underline{75.77} \\
\rowcolor[RGB]{220,228,244}
\textbf{iGSP(Ours)}&\textbf{81.14} &\textbf{77.34} &\textbf{80.55}&\textbf{76.35}&\textbf{80.01}&\textbf{76.00} \\
\bottomrule
\end{tabular}
\label{tab:cifar}
\end{table}

\begin{table}[htbp]
\caption{Comparison of state-of-the-art CL methods on CIFAR100 benchmark in class-incremental setting.}
\centering
\footnotesize
\setlength{\tabcolsep}{4pt}
\begin{tabular}{lcccccc}
\toprule
\multirow{2}{*}{Method}&
\multicolumn{2}{c}{10-step} &
\multicolumn{2}{c}{20-step} &
\multicolumn{2}{c}{50-step} \\
\cmidrule(lr){2-3} \cmidrule(lr){4-5} \cmidrule(lr){6-7}
& Avg. & Last & Avg. & Last & Avg. & Last \\
\midrule
UCIR\cite{ucir}        & 58.66 & 43.39 & 58.17 & 40.63 & 56.86 & 37.09 \\
BiC\cite{Bic}         & 68.80 & 53.54 & 66.48 & 47.02 & 62.09 & 41.04 \\
PODNet\cite{podnet}     & 58.03 & 41.05 & 53.97 & 35.02 & 51.19 & 32.99 \\
DER\cite{der}         & 74.64 & 64.35 & 73.98 & 62.55 & 72.05 & 59.76 \\
DyTox\cite{dytox}      & 74.10 & 62.34 & 71.62 & 57.43 & 68.90 & 51.09 \\
DNE\cite{dne}        & 74.86 & 70.04 & \textemdash & \textemdash & \textemdash & \textemdash \\
\midrule
CLIP zero-shot   & 74.47 & 65.92 & 75.20 & 65.74 & 75.67 & 65.94 \\
Fine-tune        & 65.46 & 53.23 & 59.69 & 43.13 & 39.23 & 18.89 \\
LwF\cite{lwf}         & 65.86 & 48.04 & 60.64 & 40.56 & 47.69 & 32.90 \\
iCaRL\cite{icarl}       & 79.35 & 70.97 & 73.32 & 64.55 & 71.28 & 59.07 \\
LwF-VR\cite{lwf-vr}      & 78.81 & 70.75 & 74.54 & 63.54 & 71.02 & 59.45 \\
ZSCL\cite{zscl}        & 82.15 & 73.65 & 80.39 & 69.58 & 79.92 & 67.36 \\
MoE-Adapter(++)\cite{moeadapter} & 85.21 & 77.52 & 83.72 & 76.20 & 83.60 & \textbf{75.24} \\
\rowcolor[RGB]{220,228,244}
\textbf{iGSP(Ours)}&\textbf{85.34} &\textbf{77.69} &\textbf{83.99}&\textbf{76.54}  &\textbf{83.82} &\underline{75.13} \\   
\bottomrule
\end{tabular}
\label{tab:tinyimage}
\end{table}

\textbf{Parameter Efficiency.} 
As illustrated in \cref{tab:train-efficiency} We further compare the training efficiency of our iGSP framework with several representative continual learning methods, iGSP achieves remarkable parameter efficiency, requiring only \textbf{0.63M} trainable parameters---about \textbf{1/250} of the full fine-tuning counterparts (149.6M). 
Moreover, iGSP maintains extremely low computational overhead, with an average GPU memory consumption of only \textbf{3.5\,GB} and a per-iteration time of \textbf{0.097\,s}. 
This demonstrates that by reusing and recombining a compact set of experts, iGSP effectively minimizes both memory footprint and computation cost, achieving the most efficient adaptation among all compared methods.
Furthermore, \cref{tab:final} compares the total parameter count and inference memory footprint after the final task. Compared to MoE-Adapters and other counterparts lacking knowledge reuse mechanisms, our method achieves a remarkable 86.9\% reduction in total parameters.

\begin{table}[htbp]
\centering
\footnotesize
\caption{Comparison of SOTA methods on training efficiency.}
\begin{tabularx}{\columnwidth}{c >{\centering\arraybackslash}X >{\centering\arraybackslash}X >{\centering\arraybackslash}X}
\toprule
Method & Train Params & GPU Mem.& Times \\
\midrule
LwF \cite{lwf}             & 149.6M & 32172\,MiB & 1.54\,s/it \\
LwF-VR\cite{lwf-vr}           & 149.6M & 32236\,MiB & 1.51\,s/it \\
ZSCL \cite{zscl}            & 149.6M & 19898\,MiB & 3.94\,s/it \\
MoE-Adapter\cite{moeadapter}      & 59.8M  & 22358\,MiB & 1.58\,s/it \\
MoE-Adapter++\cite{yu2025moe} & 1.1M & 6529\,MiB &0.68\,s/it \\ 
\rowcolor[RGB]{220,228,244} iGSP (Ours) & 0.63M & 3568\,MiB& 0.097\,s/it \\
\textcolor{red}{$\Delta$} & \textcolor{red}{-42.7\%} & \textcolor{red}{-45.4\%} & \textcolor{red}{-85.7\%} \\

\bottomrule
\end{tabularx}

\label{tab:train-efficiency}
\end{table}

\begin{table}[t]
    \centering
    \footnotesize
    \caption{Comparison of methods on final additional params.} 
    \label{tab:final}
    \begin{tabularx}{\columnwidth}{c >{\centering\arraybackslash}X}
        \toprule
        Methods & Final Additional Params (M) \\
        \midrule
        MoE-Adapters & 51.06 \\
        \rowcolor[RGB]{220,228,244} iGSP(Ours) & 6.68 \\ 
        \textcolor{red}{$\Delta$} & \textcolor{red}{-86.9\%} \\
        \bottomrule
    \end{tabularx}
\end{table}

\subsection{Ablation Study}

\begin{figure}[htbp]
  \centering
   \includegraphics[width=0.9\linewidth]{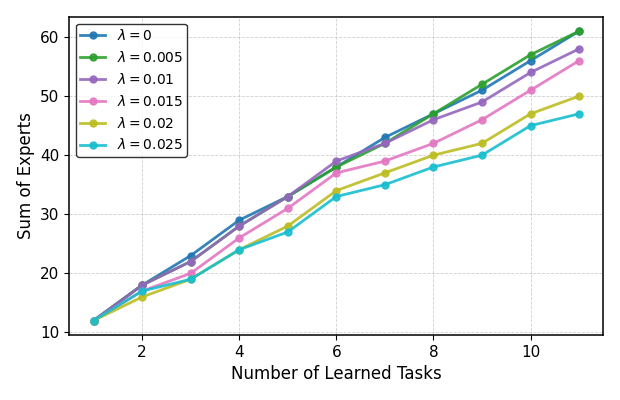}
      \caption{Number of Experts vs. Number of Learned Tasks under Different Shared Regularization Coefficient.}
   \label{fig:expertsnum}
\end{figure}

\textbf{Analysis of Subspace-Constrained Regularization.} 
To evaluate the impact of the proposed SCR, we fix the number of candidate basis vectors to $M=1$ and systematically vary the regularization coefficient $\lambda$ from $0$ to $0.025$. As illustrated in \cref{fig:expertsnum}, increasing $\lambda$ effectively dampens the growth rate of the expert population relative to the number of learned tasks. This observation confirms that a stronger SCR penalty strictly enforces gradient projection onto the historical subspace, compelling the model to exhaust the expressive capacity of existing experts before activating new orthogonal dimensions. 
Despite the constrained parameter growth, \cref{tab:lambda} demonstrates that the overall accuracy remains remarkably stable across varying $\lambda$ values. This stability suggests that our SCR formulation successfully identifies and exploits latent shared alignment structures without inducing optimization interference or performance degradation. Consequently, SCR provides a principled mechanism for achieving high adaptation efficiency by maximizing knowledge reuse within the gradient subspace.

\begin{figure}[htbp]
  \centering
   \includegraphics[width=0.9\linewidth]{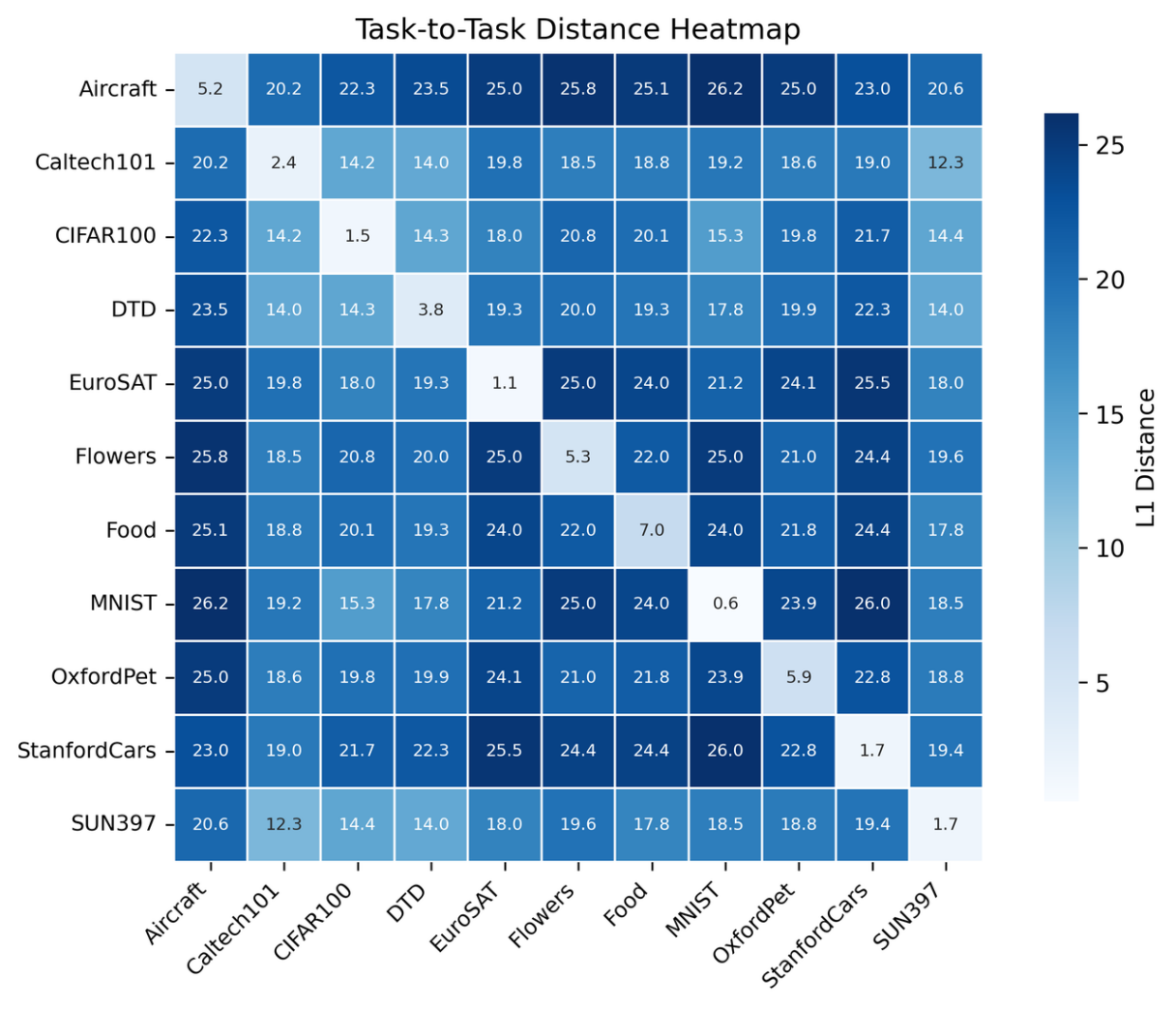}
      \caption{
  Visualization of the L1 distances between average visual-textual representations across tasks in MTIL. }
   \label{fig:taskmatrix}
\end{figure}

\textbf{Analysis of IFER Task Identification.}
To assess the robustness and discriminative capability of the proposed IFER in multi-task vision-language settings,
we examine the hidden representations of the visual and textual encoders on the MTIL benchmark.
The L1 distance between mean visual and textual embeddings is computed to quantify semantic discrepancies across tasks.
\textit{Inter-task} distances are obtained by comparing embeddings from different tasks,
while \textit{intra-task} distances are calculated between two independently sampled batches of the same task.
As shown in \cref{fig:taskmatrix}, diagonal elements with L1 distances ranging from 1.7 to 7.0 indicate strong intra-task semantic consistency and clear task separability. 

\begin{figure}[htbp]
  \centering
  \includegraphics[width=1\linewidth]{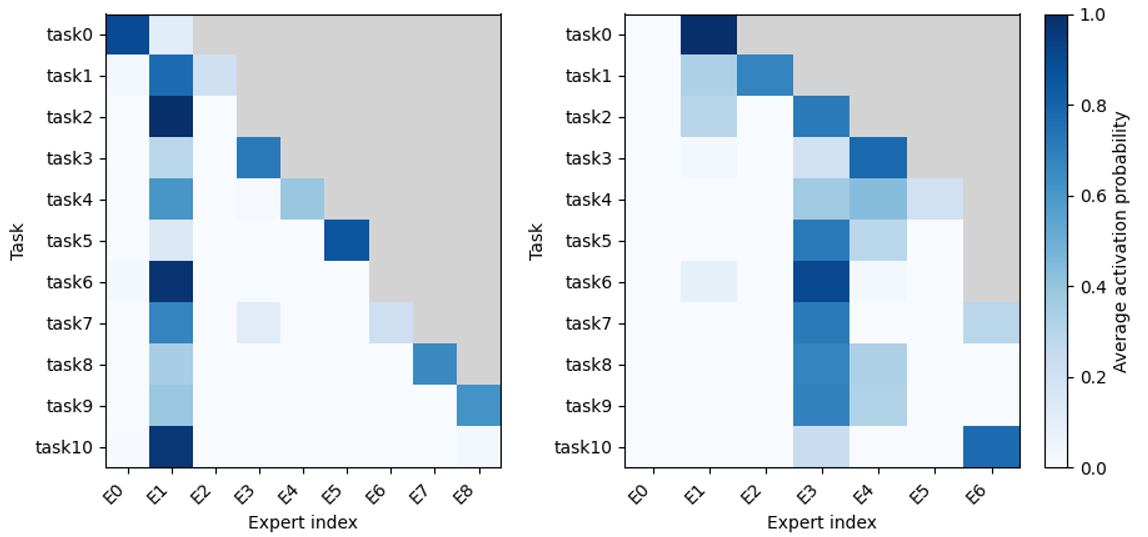}
      \caption{
  Visualization of subspace basis utilization in iGSP across tasks. 
  Left: average activation probability of experts in the 9th layer. 
  Right: 12th layer.  }
   \label{fig:expertuse}
\end{figure}

\textbf{Visualization of Subspace Basis Utilization.} 
To verify the manifestation of shared alignment in the optimization space, we visualize the average routing distribution across different tasks in \cref{fig:expertuse}. The left and right panels display the 9th and 12th layers, respectively. As theoretically derived in \cref{sec:methods}, these activation probabilities mirror the gradient flow magnitude into each subspace basis. We observe that iGSP consistently identifies a significant set of reusable basis vectors across diverse tasks during both training and inference. This high degree of geometric overlap validates our core hypothesis that visually heterogeneous tasks often converge on shared gradient subspaces for cross-modal alignment. This effect is particularly prominent in the 12th layer, where the model utilizes only 7 basis vectors to satisfy the alignment requirements of 11 complex tasks. Such extreme convergence in the deeper layers further highlights the parameter efficiency of iGSP, demonstrating its ability to capture latent shared alignment structures with a minimal set of orthogonal dimensions.
\begin{table}[htbp]
\centering
\footnotesize
\caption{Impact of $\lambda$ on model performance and trainable parameters.}
\label{tab:lambda}
\begin{tabularx}{\columnwidth}{c >{\centering\arraybackslash}X >{\centering\arraybackslash}X >{\centering\arraybackslash}X >{\centering\arraybackslash}X}
\toprule
$\lambda$ & Transfer & Average & Last & Train Params \\
\midrule
0     & 69.8 & 78.0 & 86.8 & 0.63M \\
0.005 & 69.8 & 78.0 & 86.9 & 0.63M \\
0.010 & 69.8 & 77.9 & 86.8 & 0.62M \\
0.015 & 69.8 & 78.0 & 86.8 & 0.61M \\
0.020 & 69.8 & 77.9 & 86.5 & 0.58M \\
0.025 & 69.8 & 77.8 & 86.4 & 0.56M \\
\bottomrule
\label{tab:lambda}
\end{tabularx}
\end{table}

\textbf{Impact of Subspace Pre-expansion Scale ($M$).} 
We evaluate the sensitivity of iGSP to the number of candidate basis vectors $M$ by varying it from 1 to 5 while fixing the regularization coefficient $\lambda$ at 0.05. As reported in \cref{tab:M}, the choice of $M$ has a marginal impact on the final accuracy but significantly influences the optimization dynamics during the \textit{Subspace Identification} phase. Specifically, using $M=1$ can restrict the model's representational plasticity, providing insufficient orthogonal degrees of freedom to capture complex task-specific residual alignments. Increasing $M$ to 2 enhances the subspace dimensionality, leading to more reliable fitting of novel patterns that lie beyond the historical span. 
However, beyond $M=2$, introducing further candidate basis vectors yields diminishing returns in performance while linearly increasing the number of trainable parameters during the initial optimization stage. This observation confirms that a compact set of candidates is sufficient for identifying the optimal task-specific subspace when guided by our SCR. Consistent with our theoretical framework, the computational overhead of the identification phase scales approximately linearly with $M$, justifying the use of a minimal $M$ to maintain high training efficiency without sacrificing the model's ability to discover necessary orthogonal dimensions.
\begin{table}[t]
\centering
\caption{Impact of M on model performance and trainable param-
eters.}
\footnotesize
\label{tab:M}
\begin{tabularx}{\columnwidth}{c >{\centering\arraybackslash}X >{\centering\arraybackslash}X >{\centering\arraybackslash}X >{\centering\arraybackslash}X}
\toprule
M & Last & Transfer & Average & Train Params \\
\midrule
1 & 86.85 & 69.8 & 78 & 0.63M \\
2 & 86.9  & 69.8 & 78.07 & 1.00M \\
3 & 86.64 & 69.8 & 77.86 & 1.22M \\
4 & 86.6  & 69.8 & 77.91 & 1.48M \\
5 & 86.1  & 69.8 & 77.67 & 1.75M \\
\bottomrule
\end{tabularx}
\end{table}
\section{Conclusion}
\label{sec:conclusion}
In this work, we reframe the challenge of parameter-efficient continual learning in vision-language models (VLMs) from the perspective of gradient dynamics. We observed that existing similarity-driven sharing mechanisms falsely equate superficial visual similarity with underlying alignment consistency. This fundamental mismatch inevitably triggers severe negative transfer between visually similar but logically distinct tasks, while completely failing to enable deep parameter reuse across visually diverse tasks. We argue that cross-task knowledge sharing in continual learning is fundamentally a geometric problem: tasks should share parameters if and only if their gradient updates reside in the same low-rank subspace.

To this end, we propose iGSP, a novel framework that achieves efficient adaptation via implicit gradient subspace projection. Leveraging the empirical observation that MoE routers converge early to establish the subspace basis, iGSP bifurcates the adaptation process into two optimization phases. First, in the Subspace Identification phase, a novel subspace-constrained regularization implicitly forces the gradients of new tasks to project onto the subspaces of previously learned experts. By mathematically connecting routing frequency with gradient flow magnitude, iGSP effectively prunes redundant null-space dimensions. Second, in the Orthogonal Fine-tuning phase, structural constraints are removed, allowing the isolated new orthogonal dimensions to rapidly fit the task-specific residual loss without interference.

Extensive experiments on the MTIL benchmark demonstrate that iGSP achieves state-of-the-art accuracy while reducing trainable parameters by 42.7\%. This demonstrates that iGSP efficiently utilizes computational resources while maintaining high performance for continual learning tasks in vision-language models. Moreover, iGSP offers a practical solution for deploying large-scale VLMs in resource-constrained environments such as robotics, unmanned aerial systems, and satellite-based platforms, where computational and storage limitations are critical.

Despite the significant performance improvements of iGSP, some limitations remain. Specifically, task-identity inference currently relies on a manually specified threshold for routing selection, which may affect robustness across diverse domain shifts. Future work will focus on developing more adaptive and probabilistic mechanisms for task identification and routing, further enhancing the flexibility and reliability of subspace-aware continual learning. Additionally, scaling the framework to larger VLMs to handle more complex tasks and data remains a promising area for further exploration.

Overall, the iGSP framework proposed in this study provides a novel perspective for addressing cross-task knowledge sharing in vision-language models and offers a theoretically grounded and practically viable solution for deploying large-scale continual learning systems in resource-constrained environments. We believe that with further research, iGSP will demonstrate its unique advantages and wide applicability in more real-world scenarios.



\bibliographystyle{IEEEtran}

\bibliography{refs}

@String(CVPR= {IEEE Conf. Comput. Vis. Pattern Recog.})

@String(ECCV= {Eur. Conf. Comput. Vis.})

@String(ICLR = {Int. Conf. Learn. Represent.})

@String(AAAI = {AAAI})

@String(CVPR  = {CVPR})

@String(ECCV  = {ECCV})

@String(ICLR  = {ICLR})

@inproceedings{clip,
  title={Learning transferable visual models from natural language supervision},
  author={Radford, Alec and Kim, Jong Wook and Hallacy, Chris and Ramesh, Aditya and Goh, Gabriel and Agarwal, Sandhini and Sastry, Girish and Askell, Amanda and Mishkin, Pamela and Clark, Jack and others},
  booktitle={International conference on machine learning},
  pages={8748--8763},
  year={2021},
  organization={PmLR}
}

@inproceedings{zscl,
  title={Preventing zero-shot transfer degradation in continual learning of vision-language models},
  author={Zheng, Zangwei and Ma, Mingyuan and Wang, Kai and Qin, Ziheng and Yue, Xiangyu and You, Yang},
  booktitle={Proceedings of the IEEE/CVF international conference on computer vision},
  pages={19125--19136},
  year={2023}
}

@inproceedings{icarl,
  title={icarl: Incremental classifier and representation learning},
  author={Rebuffi, Sylvestre-Alvise and Kolesnikov, Alexander and Sperl, Georg and Lampert, Christoph H},
  booktitle={Proceedings of the IEEE conference on Computer Vision and Pattern Recognition},
  pages={2001--2010},
  year={2017}
}

@article{lwf,
  title={Learning without forgetting},
  author={Li, Zhizhong and Hoiem, Derek},
  journal={IEEE transactions on pattern analysis and machine intelligence},
  volume={40},
  number={12},
  pages={2935--2947},
  year={2017},
  publisher={IEEE}
}

@inproceedings{diki,
  title={Mind the interference: Retaining pre-trained knowledge in parameter efficient continual learning of vision-language models},
  author={Tang, Longxiang and Tian, Zhuotao and Li, Kai and He, Chunming and Zhou, Hantao and Zhao, Hengshuang and Li, Xiu and Jia, Jiaya},
  booktitle={European conference on computer vision},
  pages={346--365},
  year={2024},
  organization={Springer}
}

@inproceedings{l2p,
  title={Learning to prompt for continual learning},
  author={Wang, Zifeng and Zhang, Zizhao and Lee, Chen-Yu and Zhang, Han and Sun, Ruoxi and Ren, Xiaoqi and Su, Guolong and Perot, Vincent and Dy, Jennifer and Pfister, Tomas},
  booktitle={Proceedings of the IEEE/CVF conference on computer vision and pattern recognition},
  pages={139--149},
  year={2022}
}

@inproceedings{dualprompt,
  title={Dualprompt: Complementary prompting for rehearsal-free continual learning},
  author={Wang, Zifeng and Zhang, Zizhao and Ebrahimi, Sayna and Sun, Ruoxi and Zhang, Han and Lee, Chen-Yu and Ren, Xiaoqi and Su, Guolong and Perot, Vincent and Dy, Jennifer and others},
  booktitle={European conference on computer vision},
  pages={631--648},
  year={2022},
  organization={Springer}
}

@article{s-prompts,
  title={S-prompts learning with pre-trained transformers: An occam’s razor for domain incremental learning},
  author={Wang, Yabin and Huang, Zhiwu and Hong, Xiaopeng},
  journal={Advances in Neural Information Processing Systems},
  volume={35},
  pages={5682--5695},
  year={2022}
}

@inproceedings{moeadapter,
  title={Boosting continual learning of vision-language models via mixture-of-experts adapters},
  author={Yu, Jiazuo and Zhuge, Yunzhi and Zhang, Lu and Hu, Ping and Wang, Dong and Lu, Huchuan and He, You},
  booktitle={Proceedings of the IEEE/CVF Conference on Computer Vision and Pattern Recognition},
  pages={23219--23230},
  year={2024}
}

@article{lwf-vr,
  title={Don't stop learning: Towards continual learning for the clip model},
  author={Ding, Yuxuan and Liu, Lingqiao and Tian, Chunna and Yang, Jingyuan and Ding, Haoxuan},
  journal={arXiv preprint arXiv:2207.09248},
  year={2022}
}

@article{ewc,
  title={Overcoming catastrophic forgetting in neural networks},
  author={Kirkpatrick, James and Pascanu, Razvan and Rabinowitz, Neil and Veness, Joel and Desjardins, Guillaume and Rusu, Andrei A and Milan, Kieran and Quan, John and Ramalho, Tiago and Grabska-Barwinska, Agnieszka and others},
  journal={Proceedings of the national academy of sciences},
  volume={114},
  number={13},
  pages={3521--3526},
  year={2017},
  publisher={National Academy of Sciences}
}

@article{vit,
  title={An image is worth 16x16 words: Transformers for image recognition at scale},
  author={Dosovitskiy, Alexey},
  journal={arXiv preprint arXiv:2010.11929},
  year={2020}
}

@article{lora,
  title={Lora: Low-rank adaptation of large language models.},
  author={Hu, Edward J and Shen, Yelong and Wallis, Phillip and Allen-Zhu, Zeyuan and Li, Yuanzhi and Wang, Shean and Wang, Lu and Chen, Weizhu and others},
  journal={ICLR},
  volume={1},
  number={2},
  pages={3},
  year={2022}
}

@article{llava,
  title={Visual instruction tuning},
  author={Liu, Haotian and Li, Chunyuan and Wu, Qingyang and Lee, Yong Jae},
  journal={Advances in neural information processing systems},
  volume={36},
  pages={34892--34916},
  year={2023}
}

@inproceedings{inflora,
  title={Inflora: Interference-free low-rank adaptation for continual learning},
  author={Liang, Yan-Shuo and Li, Wu-Jun},
  booktitle={Proceedings of the IEEE/CVF Conference on Computer Vision and Pattern Recognition},
  pages={23638--23647},
  year={2024}
}

@inproceedings{sema,
  title={Self-expansion of pre-trained models with mixture of adapters for continual learning},
  author={Wang, Huiyi and others},
  booktitle={CVPR},
  pages={10087--10098},
  year={2025}
}

@inproceedings{wise-ft,
  title={Robust fine-tuning of zero-shot models},
  author={Wortsman, Mitchell and Ilharco, Gabriel and Kim, Jong Wook and Li, Mike and Kornblith, Simon and Roelofs, Rebecca and Lopes, Raphael Gontijo and Hajishirzi, Hannaneh and Farhadi, Ali and Namkoong, Hongseok and others},
  booktitle={Proceedings of the IEEE/CVF conference on computer vision and pattern recognition},
  pages={7959--7971},
  year={2022}
}

@article{cifar100,
  title={Learning multiple layers of features from tiny images},
  author={Krizhevsky, Alex and Hinton, Geoffrey and others},
  year={2009},
  publisher={Toronto, ON, Canada}
}

@inproceedings{dtd,
  title={Describing textures in the wild},
  author={Cimpoi, Mircea and Maji, Subhransu and Kokkinos, Iasonas and Mohamed, Sammy and Vedaldi, Andrea},
  booktitle={Proceedings of the IEEE conference on computer vision and pattern recognition},
  pages={3606--3613},
  year={2014}
}

@inproceedings{gift,
  title={Synthetic Data is an Elegant GIFT for Continual Vision-Language Models},
  author={Wu, Bin and Shi, Wuxuan and Wang, Jinqiao and Ye, Mang},
  booktitle={Proceedings of the Computer Vision and Pattern Recognition Conference},
  pages={2813--2823},
  year={2025}
}

@article{eurosat,
  title={Eurosat: A novel dataset and deep learning benchmark for land use and land cover classification},
  author={Helber, Patrick and Bischke, Benjamin and Dengel, Andreas and Borth, Damian},
  journal={IEEE Journal of Selected Topics in Applied Earth Observations and Remote Sensing},
  volume={12},
  number={7},
  pages={2217--2226},
  year={2019},
  publisher={IEEE}
}

@inproceedings{flowers,
  title={Automated flower classification over a large number of classes},
  author={Nilsback, Maria-Elena and Zisserman, Andrew},
  booktitle={2008 Sixth Indian conference on computer vision, graphics \& image processing},
  pages={722--729},
  year={2008},
  organization={IEEE}
}

@inproceedings{food,
  title={Food-101--mining discriminative components with random forests},
  author={Bossard, Lukas and Guillaumin, Matthieu and Van Gool, Luc},
  booktitle={European conference on computer vision},
  pages={446--461},
  year={2014},
  organization={Springer}
}

@article{mnist,
  title={The mnist database of handwritten digit images for machine learning research [best of the web]},
  author={Deng, Li},
  journal={IEEE signal processing magazine},
  volume={29},
  number={6},
  pages={141--142},
  year={2012},
  publisher={IEEE}
}

@article{aircraft,
  title={Fine-grained visual classification of aircraft},
  author={Maji, Subhransu and Rahtu, Esa and Kannala, Juho and Blaschko, Matthew and Vedaldi, Andrea},
  journal={arXiv preprint arXiv:1306.5151},
  year={2013}
}

@inproceedings{ucir,
  title={Learning a unified classifier incrementally via rebalancing},
  author={Hou, Saihui and Pan, Xinyu and Loy, Chen Change and Wang, Zilei and Lin, Dahua},
  booktitle={Proceedings of the IEEE/CVF conference on computer vision and pattern recognition},
  pages={831--839},
  year={2019}
}

@inproceedings{Bic,
  title={Large scale incremental learning},
  author={Wu, Yue and Chen, Yinpeng and Wang, Lijuan and Ye, Yuancheng and Liu, Zicheng and Guo, Yandong and Fu, Yun},
  booktitle={Proceedings of the IEEE/CVF conference on computer vision and pattern recognition},
  pages={374--382},
  year={2019}
}

@inproceedings{podnet,
  title={Podnet: Pooled outputs distillation for small-tasks incremental learning},
  author={Douillard, Arthur and Cord, Matthieu and Ollion, Charles and Robert, Thomas and Valle, Eduardo},
  booktitle={European Conference on Computer Vision},
  pages={86--102},
  year={2020},
  organization={Springer}
}

@inproceedings{der,
  title={Der: Dynamically expandable representation for class incremental learning},
  author={Yan, Shipeng and Xie, Jiangwei and He, Xuming},
  booktitle={Proceedings of the IEEE/CVF conference on computer vision and pattern recognition},
  pages={3014--3023},
  year={2021}
}

@inproceedings{dytox,
  title={Dytox: Transformers for continual learning with dynamic token expansion},
  author={Douillard, Arthur and Ram{\'e}, Alexandre and Couairon, Guillaume and Cord, Matthieu},
  booktitle={Proceedings of the IEEE/CVF conference on computer vision and pattern recognition},
  pages={9285--9295},
  year={2022}
}

@inproceedings{dne,
  title={Dense network expansion for class incremental learning},
  author={Hu, Zhiyuan and Li, Yunsheng and Lyu, Jiancheng and Gao, Dashan and Vasconcelos, Nuno},
  booktitle={Proceedings of the IEEE/CVF Conference on Computer Vision and Pattern Recognition},
  pages={11858--11867},
  year={2023}
}

@inproceedings{eeil,
  title={End-to-end incremental learning},
  author={Castro, Francisco M and Mar{\'\i}n-Jim{\'e}nez, Manuel J and Guil, Nicol{\'a}s and Schmid, Cordelia and Alahari, Karteek},
  booktitle={Proceedings of the European conference on computer vision (ECCV)},
  pages={233--248},
  year={2018}
}

@inproceedings{muc,
  title={More classifiers, less forgetting: A generic multi-classifier paradigm for incremental learning},
  author={Liu, Yu and Parisot, Sarah and Slabaugh, Gregory and Jia, Xu and Leonardis, Ales and Tuytelaars, Tinne},
  booktitle={European Conference on Computer Vision},
  pages={699--716},
  year={2020},
  organization={Springer}
}

@inproceedings{pass,
  title={Prototype augmentation and self-supervision for incremental learning},
  author={Zhu, Fei and Zhang, Xu-Yao and Wang, Chuang and Yin, Fei and Liu, Cheng-Lin},
  booktitle={Proceedings of the IEEE/CVF conference on computer vision and pattern recognition},
  pages={5871--5880},
  year={2021}
}

@article{prefix,
  title={Prefix-tuning: Optimizing continuous prompts for generation},
  author={Li, Xiang Lisa and Liang, Percy},
  journal={arXiv preprint arXiv:2101.00190},
  year={2021}
}

@inproceedings{replay1,
  title={Selective experience replay for lifelong learning},
  author={Isele, David and Cosgun, Akansel},
  booktitle={Proceedings of the AAAI conference on artificial intelligence},
  volume={32},
  number={1},
  year={2018}
}

@inproceedings{re1,
  title={Memory aware synapses: Learning what (not) to forget},
  author={Aljundi, Rahaf and Babiloni, Francesca and Elhoseiny, Mohamed and Rohrbach, Marcus and Tuytelaars, Tinne},
  booktitle={Proceedings of the European conference on computer vision (ECCV)},
  pages={139--154},
  year={2018}
}

@article{moe,
  title={Adaptive mixtures of local experts},
  author={Jacobs, Robert A and Jordan, Michael I and Nowlan, Steven J and Hinton, Geoffrey E},
  journal={Neural computation},
  volume={3},
  number={1},
  pages={79--87},
  year={1991},
  publisher={MIT Press}
}

@inproceedings{clmoe,
  title={CL-MoE: Enhancing Multimodal Large Language Model with Dual Momentum Mixture-of-Experts for Continual Visual Question Answering},
  author={Huai, Tianyu and Zhou, Jie and Wu, Xingjiao and Chen, Qin and Bai, Qingchun and Zhou, Ze and He, Liang},
  booktitle={Proceedings of the Computer Vision and Pattern Recognition Conference},
  pages={19608--19617},
  year={2025}
}

@incollection{catastrophic,
  title={Catastrophic interference in connectionist networks: The sequential learning problem},
  author={McCloskey, Michael and Cohen, Neal J},
  booktitle={Psychology of learning and motivation},
  volume={24},
  pages={109--165},
  year={1989},
  publisher={Elsevier}
}

@article{continualsurvey,
  title={Continual Learning for VLMs: A Survey and Taxonomy Beyond Forgetting},
  author={Liu, Yuyang and Hong, Qiuhe and Huang, Linlan and Gomez-Villa, Alexandra and Goswami, Dipam and Liu, Xialei and van de Weijer, Joost and Tian, Yonghong},
  journal={arXiv preprint arXiv:2508.04227},
  year={2025}
}

@inproceedings{cllora,
  title={CL-LoRA: Continual Low-Rank Adaptation for Rehearsal-Free Class-Incremental Learning},
  author={He, Jiangpeng and Duan, Zhihao and Zhu, Fengqing},
  booktitle={Proceedings of the Computer Vision and Pattern Recognition Conference},
  pages={30534--30544},
  year={2025}
}

@article{adamW,
  title={Decoupled weight decay regularization},
  author={Loshchilov, Ilya and Hutter, Frank},
  journal={arXiv preprint arXiv:1711.05101},
  year={2017}
}

@article{ptmcontinual,
  title={Continual learning with pre-trained models: A survey},
  author={Zhou, Da-Wei and Sun, Hai-Long and Ning, Jingyi and Ye, Han-Jia and Zhan, De-Chuan},
  journal={arXiv preprint arXiv:2401.16386},
  year={2024}
}

@inproceedings{aligment,
  title={Assessing and Learning Alignment of Unimodal Vision and Language Models},
  author={Zhang, Le and Yang, Qian and Agrawal, Aishwarya},
  booktitle={Proceedings of the Computer Vision and Pattern Recognition Conference},
  pages={14604--14614},
  year={2025}
}

@article{pecl1,
  title={Class incremental learning with pre-trained vision-language models},
  author={Liu, Xialei and Cao, Xusheng and Lu, Haori and Xiao, Jia-wen and Bagdanov, Andrew D and Cheng, Ming-Ming},
  journal={arXiv preprint arXiv:2310.20348},
  year={2023}
}

@inproceedings{bilora,
  title={BiLoRA: Almost-Orthogonal Parameter Spaces for Continual Learning},
  author={Zhu, Hao and Zhang, Yifei and Dong, Junhao and Koniusz, Piotr},
  booktitle={Proceedings of the Computer Vision and Pattern Recognition Conference},
  pages={25613--25622},
  year={2025}
}

@inproceedings{yu2025language,
  title={Language Guided Concept Bottleneck Models for Interpretable Continual Learning},
  author={Yu, Lu and Han, Haoyu and Tao, Zhe and Yao, Hantao and Xu, Changsheng},
  booktitle={Proceedings of the Computer Vision and Pattern Recognition Conference},
  pages={14976--14986},
  year={2025}
}

@inproceedings{kang2025your,
  title={Do Your Best and Get Enough Rest for Continual Learning},
  author={Kang, Hankyul and Seifer, Gregor and Lee, Donghyun and Ryu, Jongbin},
  booktitle={Proceedings of the Computer Vision and Pattern Recognition Conference},
  pages={10077--10086},
  year={2025}
}

@article{yu2025moe,
  title={MoE-Adapters++: Towards More Efficient Continual Learning of Vision-Language Models via Dynamic Mixture-of-Experts Adapters},
  author={Yu, Jiazuo and Huang, Zichen and Zhuge, Yunzhi and Zhang, Lu and Hu, Ping and Wang, Dong and Lu, Huchuan and He, You},
  journal={IEEE Transactions on Pattern Analysis and Machine Intelligence},
  year={2025},
  publisher={IEEE}
}

@ARTICLE{squeezing,
  author={Yu, Da and Zhang, Mingyi and Li, Mantian and Zha, Fusheng and Zhang, Junge and Sun, Lining and Huang, Kaiqi},
  journal={IEEE/CAA Journal of Automatica Sinica}, 
  title={Squeezing More Past Knowledge for Online Class-Incremental Continual Learning}, 
  year={2023},
  volume={10},
  number={3},
  pages={722-736},
  doi={10.1109/JAS.2023.123090}}

@article{gem,
  title={Gradient episodic memory for continual learning},
  author={Lopez-Paz, David and Ranzato, Marc'Aurelio},
  journal={Advances in neural information processing systems},
  volume={30},
  year={2017}
}

@article{ogd,
  title={Understanding and improving information transfer in multi-task learning},
  author={Wu, Sen and Zhang, Hongyang R and R{\'e}, Christopher},
  journal={arXiv preprint arXiv:2005.00944},
  year={2020}
}

@inproceedings{yang2023data,
  title={Data augmented flatness-aware gradient projection for continual learning},
  author={Yang, Enneng and Shen, Li and Wang, Zhenyi and Liu, Shiwei and Guo, Guibing and Wang, Xingwei},
  booktitle={Proceedings of the IEEE/CVF international conference on computer vision},
  pages={5630--5639},
  year={2023}
}

@inproceedings{zhao2023rethinking,
  title={Rethinking gradient projection continual learning: Stability/plasticity feature space decoupling},
  author={Zhao, Zhen and Zhang, Zhizhong and Tan, Xin and Liu, Jun and Qu, Yanyun and Xie, Yuan and Ma, Lizhuang},
  booktitle={Proceedings of the IEEE/CVF conference on computer vision and pattern recognition},
  pages={3718--3727},
  year={2023}
}

@inproceedings{apolinario2025code,
  title={Code-cl: Conceptor-based gradient projection for deep continual learning},
  author={Apolinario, Marco PE and Choudhary, Sakshi and Roy, Kaushik},
  booktitle={Proceedings of the IEEE/CVF International Conference on Computer Vision},
  pages={775--784},
  year={2025}
}

@article{lu2024visual,
  title={Visual prompt tuning in null space for continual learning},
  author={Lu, Yue and Zhang, Shizhou and Cheng, De and Xing, Yinghui and Wang, Nannan and Wang, Peng and Zhang, Yanning},
  journal={Advances in neural information processing systems},
  volume={37},
  pages={7878--7901},
  year={2024}
}

@inproceedings{qiao2024prompt,
  title={Prompt gradient projection for continual learning},
  author={Qiao, Jingyang and Tan, Xin and Chen, Chengwei and Qu, Yanyun and Peng, Yong and Xie, Yuan and others},
  booktitle={The Twelfth International Conference on Learning Representations},
  year={2024}
}

@article{luo2026keeplora,
  title={KeepLoRA: Continual Learning with Residual Gradient Adaptation},
  author={Luo, Mao-Lin and Zhou, Zi-Hao and Zhang, Yi-Lin and Wan, Yuanyu and Wei, Tong and Zhang, Min-Ling},
  journal={arXiv preprint arXiv:2601.19659},
  year={2026}
}

@article{qiu2025splitlora,
  title={SplitLoRA: Balancing Stability and Plasticity in Continual Learning Through Gradient Space Splitting},
  author={Qiu, Haomiao and Zhang, Miao and Qiao, Ziyue and Guan, Weili and Zhang, Min and Nie, Liqiang},
  journal={arXiv preprint arXiv:2505.22370},
  year={2025}
}

@article{peng2025gnsp,
  title={GNSP: Gradient Null Space Projection for Preserving Cross-Modal Alignment in VLMs Continual Learning},
  author={Peng, Tiantian and Liu, Yuyang and Yang, Shuo and Hong, Qiuhe and Tian, YongHong},
  journal={arXiv preprint arXiv:2507.19839},
  year={2025}
}

@inproceedings{kang2025dynamic,
  title={Dynamic multi-layer null space projection for vision-language continual learning},
  author={Kang, Borui and Wang, Lei and Wu, Zhiping and Feng, Tao and Li, Yawen and Gao, Yang and Li, Wenbin},
  booktitle={Proceedings of the IEEE/CVF International Conference on Computer Vision},
  pages={2077--2086},
  year={2025}
}

@ARTICLE{zhang2024continual,
  author={Zhang, Wentao and Huang, Yujun and Zhang, Weizhuo and Zhang, Tong and Lao, Qicheng and Yu, Yue and Zheng, Wei-Shi and Wang, Ruixuan},
  journal={IEEE Transactions on Circuits and Systems for Video Technology}, 
  title={Continual Learning of Image Classes With Language Guidance From a Vision-Language Model}, 
  year={2024},
  volume={34},
  number={12},
  pages={13152-13163},
  doi={10.1109/TCSVT.2024.3449109}}

@ARTICLE{wu2024image,
  author={Wu, Chenhao and Wu, Qingbo and Ma, Rui and Ngan, King Ngi and Li, Hongliang and Meng, Fanman and Qiu, Heqian},
  journal={IEEE Transactions on Circuits and Systems for Video Technology}, 
  title={Continual Cross-Domain Image Compression via Entropy Prior Guided Knowledge Distillation and Scalable Decoding}, 
  year={2024},
  volume={34},
  number={9},
  pages={8080-8092},
  doi={10.1109/TCSVT.2024.3385444}}

@ARTICLE{xiong2026class,
  author={Xiong, Fangying and Yuan, Zhaoquan and Wu, Xiao and Xu, Changsheng},
  journal={IEEE Transactions on Circuits and Systems for Video Technology}, 
  title={Class-Specific Knowledge-Guided Multimodal Prompt Tuning for Few-Shot Class-Incremental Learning}, 
  year={2026},
  volume={36},
  number={1},
  pages={763-776},
  doi={10.1109/TCSVT.2025.3597447}}

@ARTICLE{feng2025pectp,
  author={Feng, Qian and Zhao, Hanbin and Zhang, Chao and Dong, Jiahua and Ding, Henghui and Jiang, Yu-Gang and Qian, Hui},
  journal={IEEE Transactions on Circuits and Systems for Video Technology}, 
  title={PECTP: Parameter-Efficient Cross-Task Prompts for Incremental Vision Transformer}, 
  year={2025},
  volume={35},
  number={11},
  pages={11282-11296},
  doi={10.1109/TCSVT.2025.3572943}}

\newpage

\vfill

\end{document}